\documentclass[11pt]{article}

\usepackage{tabularx}
\usepackage[all]{xy}
\usepackage[text={6.5in,9.0in},centering]{geometry}

\usepackage{amsfonts}
\usepackage{amsmath}
\usepackage{amssymb} 
\usepackage{url}
\usepackage{subfigure}
\usepackage{psfrag}
\usepackage{url}
\usepackage{setspace}
\usepackage{endnotes}
\usepackage{harvard}
\usepackage[pdftex]{graphicx}
\usepackage[linesnumbered]{algorithm2e}

\newcommand{\Pa}{\operatorname{Pa}}

\newcommand{\argmax}{\operatornamewithlimits{arg\ max}}
\newcommand{\argmin}{\operatornamewithlimits{arg\ min}}

\newcommand{\var}{\operatorname{Var}}

\title{Distributed Constrained Optimization with Semicoordinate
Transformations}

\author{William Macready (\href{mailto:wgm@dwavesys.com}{\texttt{wgm@dwavesys.com}}) \\ \\
D-Wave Systems \\
Suite 100, 4401 Still Creek Dr \\ Burnaby, BC, V5C 6G9
\\ \\ David Wolpert (\href{mailto:dhw@email.arc.nasa.gov}{\texttt{dhw@email.arc.nasa.gov}}) \\ \\
NASA Ames
Research Center, \\ MailStop 269-2,\\
Moffett Field, CA, 94035 }

\begin{document}

\maketitle

\begin{abstract} Recent work has shown
how information theory extends conventional full-rationality game
theory to allow bounded rational agents. The associated
mathematical framework can be used to solve constrained
optimization problems.  This is done by translating the problem
into an iterated game, where each agent controls a different
variable of the problem, so that the joint probability
distribution across the agents' moves gives an expected value of
the objective function. The dynamics of the agents is designed to
minimize a Lagrangian function of that joint distribution. Here we
illustrate how the updating of the Lagrange parameters in the
Lagrangian is a form of automated annealing, which focuses the
joint distribution more and more tightly about the joint moves
that optimize the objective function.  We then investigate the use
of ``semicoordinate'' variable transformations. These separate the
joint state of the agents from the variables of the optimization
problem, with the two connected by an onto mapping. We present
experiments illustrating the ability of such transformations to
facilitate optimization. We focus on the special kind of
transformation in which the statistically independent states of
the agents induces a mixture distribution over the optimization
variables. Computer experiment illustrate this for $k$-sat
constraint satisfaction problems and for unconstrained
minimization of $NK$ functions.
\end{abstract}

\noindent \textit{Subject Classification:} programming: nonlinear,
algorithms, theory; probability: applications

\noindent \textit{Area of Review:} optimization

\nocite{bodo00}

\newpage

\section{Introduction} \label{sec:intro}

\subsection{Distributed optimization and control with Probability
Collectives}

As first described in \cite{wolp03b,wolp04a}, it turns out
that one can translate many of the concepts from statistical
physics, game theory, distributed optimization and distributed
control into one another. This translation is based on the fact that
those concepts all involve distributed systems in which the random
variables are, at any single instant, statistically independent.
(What is coupled is instead the distributions of those variables.)
Using this translation, one can transfer theory and techniques
between those fields, creating a large common mathematics that
connects them.  This common mathematics is known as Probability
Collectives (PC). Its unifying concern is the set of probability
distributions that govern any particular distributed system, and how
to manipulate those distributions to optimize one or more objective
functions. See \cite{wotu03a,wotu01a} for earlier, less formal work
on this topic.

In this paper we consider the use of PC to solve constrained
optimization and/or control problems. Reflecting the focus of PC on
distributed systems, its use for such problems is particularly
appropriate when the variables in the collective are spread across
many physically separated agents with limited inter-agent
communication (e.g., in a distributed design or supply chain
application, or distributed control). A general advantage of PC for
such problems is that since they work with probabilities rather than
the underlying variables, they can be implemented for arbitrary
types of the underlying variables. This same characteristic also
means they provides multiple solutions, each of which is robust,
along with sensitivity information concerning those solutions.  An
advantage particulary relevant to optimization is that the
distributed PC algorithm can often be implemented on a parallel
computer. An advantage particularly relevant to control problems is
that PC algorithms can, if desired, be used without any modelling
assumptions about the (stochastic) system being controlled. These
advantages are discussed in more detail below.

\subsection{The Probability Collectives Approach}

Broadly speaking, the PC approach to optimization/control is as
follows. First one maps the provided problem into a multi-agent
collective. In the simplest version of this process one assigns a
separate agent of the collective to determine the value of each of
the variables $x_i \in \mathcal{X}_i$ in the problem that we
control. So for example if  the $i$'th variable can only take on a
finite number of values, those $|\mathcal{X}_i|$ possible values
constitute the possible moves of the $i$'th
agent.\endnote{$|\mathcal{S}|$ denotes the number of elements in
the set $\mathcal{S}$.} The value of the joint set of $n$
variables (agents) describing the system is then
$\mathbf{x}=[x_1,\cdots, x_n] \in \mathcal{X}$ with $\mathcal{X}
\triangleq \mathcal{X}_1 \times \cdots \times
\mathcal{X}_n$.\endnote{In this paper vectors are indicated in
bold font and scalars are in regular font.}

Unlike many optimization methods, in PC the variables are not
manipulated directly. Rather a probability distribution is what is
manipulated. To avoid combinatorial explosions as the number of
dimensions of $\mathcal{X}$ grows, we must restrict attention to a
low-dimensional subset of the space of all probability
distributions. We indicate this by writing our distributions as $q
\in \cal{Q}$ over $\mathcal{X}$. The manipulation of that $q$
proceeds through an iterative process. The ultimate goal of this
process is to induce a distribution that is highly peaked about the
$\mathbf{x}$ optimizing the objective function $G(\mathbf{x})$,
sometimes called the {\emph{world cost}} or {\emph{world utility}}
function. (In this paper we only consider problems with a single
overall objective function, and we arbitrarily choose lower values
to be better, even when using the term ``utility''.)

In the precise algorithms investigated here, at the start of any
iteration a single Lagrangian function of $q$,
$\mathcal{L}:\mathcal{Q}\rightarrow \mathbb{R}$, is specified,
based on $G(\mathbf{x})$ and the associated constraints of the
optimization problem. Rather than minimize the objective function
over the space $\mathcal{X}$, the algorithm minimizes that
Lagrangian over $q \in \mathcal{Q}$. This is done by direct
manipulation of the components of $q$ by the agents.

After such a minimization of a Lagrangian, one modifies the
Lagrangian slightly. This is done so that the $q$ optimizing the new
Lagrangian is more tightly concentrated about $\mathbf{x}$ that
solve our optimization problem than is the current $q$. One then
uses the current $q$ as the starting point for another process of
having the agents minimize a Lagrangian, this time having them work
on that new Lagrangian.

At the end of a sequence of such iterations one ends up with a final
$q$. That $q$ is then used to determine a final answer in
$\mathcal{X}$, e.g., by sampling $q$, evaluating its mode,
evaluating its mean (if that is defined), etc. For a properly chosen
sequence of Lagrangians and algorithm for minimizing the
Lagrangians, this last step should, with high probability, provide
the desired optimal point in $\mathcal{X}$.

For the class of Lagrangians used in this paper, the sequence of
minimizations of Lagrangians is closely related to simulated
annealing.  The difference is that in simulated annealing an
inefficient Metropolis sampling process is used to implicitly
descend each iteration's Lagrangian. By explicitly manipulating $q$,
PC allows for more efficient descent.

In this paper we shall consider the case where $\cal{Q}$ is a
product space, $q(\mathbf{x}) = \prod_i q_i(x_i)$. The associated
formulation of PC is sometimes called ``Product Distribution''
theory. It corresponds to noncooperative game theory, with each
$q_i$ being agent $i$'s ``mixed strategy'' \cite{wolp04a,futi91}.
Our particular focus is the use of such product distributions when
$\mathcal{X}$ is not the same as the ultimate space of the
optimization variables, $\cal{Z}$. In this formulation --- a
modification of what was presented above --- there is an
intermediate mapping from $\cal{X} \rightarrow \cal{Z}$, and the
provided $G$ is actually a function over $\cal{Z}$, not (directly)
over $\cal{X}$. Such intermediate mappings are called ``semicoordinate
systems'', and going from one to another is a ``semicoordinate
transformation''. As elaborated below, such transformations allow
arbitrary coupling among the variables in $\cal{Z}$ while preserving
many of the computational advantages of using product distributions
over $\cal{X}$.

\subsection{Advantages of Probability Collectives}

There are many advantages to working with distribution in
$\mathcal{Q}$ rather than points in $\mathcal{X}$. Usually the
support of $q$ is all of $\mathcal{X}$, i.e., the $q$ minimizing the
Lagrangian lies in the interior of the unit simplices giving
$\mathcal{Q}$. Conversely, any element of $\mathcal{X}$ can be
viewed as a probability distribution on the edge (a vertex) of those
simplices. So working with $\mathcal{X}$ is a special case of
working with $\mathcal{Q}$, where one sticks to the vertices of
$\cal{Q}$. In this, optimizing over $\mathcal{Q}$ rather than
$\mathcal{X}$ is analogous to interior point methods. Due to the
breadth of the support of $q$, minimizing over it can also be viewed
as a way to allow information from the value of the objective
function at all ${\bf{x}} \in {\mathcal{X}}$ to be exploited
simultaneously.

Another advantage, alluded to above, is that by working with
distributions $\mathcal{Q}$ rather than the space $\mathcal{X}$,
the same general PC approach can be used for essentially any
$\mathcal{X}$, be it continuous, discrete, time-extended, mixtures
of these, etc. (Formally, those different spaces just correspond
to different probability measures, as far as PC is concerned.) For
expository simplicity though, here we will work with finite
$\mathcal{X}$, and therefore have probability distributions rather
than density functions, sums rather than integrals, etc. See in
particular \cite{biwo04a,wobi04b,wolp04g,wost06} for analysis
explicitly for the case of infinite $\mathcal{X}$.

Yet another advantage arises from the fact that when $\mathcal{X}$
is finite, $q\in\mathcal{Q}$ is a vector in a Euclidean space.
Accordingly the Lagrangian we are minimizing is a real-valued
function of a Euclidean vector. This means PC allows us to
leverage the power of descent schemes for continuous spaces like
gradient descent or Newton's method --- even if $\mathcal{X}$ is a
categorical, finite space. So with PC, schemes like ``gradient
descent for categorical variables'' are perfectly well-defined.

While the Lagrangians can be based on prior knowledge or modelling
assumptions concerning the problem, they need not be.  Nor does
optimization of a Lagrangian require control of all variables
$\mathcal{X}$ (i.e., some of the variables can be noisy). This
allows PC to be very broadly applicable.

\subsection{Connection with other sciences}

A more general advantage of PC is how it relates seemingly disparate
disciplines to one another. In particular, it can be motivated by
using information theory to relate bounded rational game theory to
statistical physics~\cite{wolp03b,wolp04a}. This allows techniques
from one field to be imported into the other field. For example, as
illustrated below, the grand canonical ensemble of physics can be
imported into noncooperative game theory to analyze games having
stochastic numbers of the players of various types.

To review, a noncooperative game consists of a sequence of stages. At
the beginning of each stage every agent (aka ``player'') sets a
probability distribution (its ``mixed strategy'') over its
moves~\cite{futi91,auha92,baol99,fule98}. The joint move at the stage
is then formed by agents simultaneously sampling their mixed
strategies at that stage. So the moves those agents make at any
particular stage of the game are statistically independent and the
distribution of the joint-moves at any stage is a product distribution
--- just like in PC theory.

This does not mean that the moves of the agents across all time are
statistically independent however. At each stage of the game each
agent will set its mixed strategy based on information gleaned from
preceding stages, information that in general will reflect the earlier
moves of the other agents. So the agents are coupled indirectly,
across time, via the updating of the $\{q_i\}_{i=1}^n$ at the end of
each stage.

Analogously, consider again the iterative PC algorithm outlined
above, and in particular the process of optimizing the Lagrangian
within some particular single iteration. Typically that process
proceeds by successively modifying $q$ across a sequence of
timesteps.  In each of those timesteps $q({\mathbf{x}}) = \prod_i
q_i({x}_i)$ is first sampled, and then it is updated based on all
previous samples. So just like in a noncooperative game there is
no direct coupling of the values of the underlying variables
\{${x}_i$\}at any particular timestep ($q$ is a product
distribution). Rather just like in a noncooperative game, the
variables are indirectly coupled, across time (i.e., across
timesteps of the optimization), via coupling of the distributions
$q_i(x_i)$ at different timesteps.

In addition, information theory can be used to show that the
bounded rational equilibrium of any noncooperative game is the $q$
optimizing an associated ``maxent Lagrangian'' $\mathcal{L}(q)$
~\cite{wolp04a}. (That Lagrangian is minimized by the distribution
that has maximal entropy while being consistent with specified
values of the average payoffs of the agents.)  This Lagrangian
turns out to be exactly the one that arises in the version of PC
considered in this paper. So bounded rational game theory is an
instance of PC.

Now in statistical physics often one wishes to find the
distribution out of an allowed set of distributions (e.g.,
$\mathcal{Q}$) with minimal distance to a fixed target
distribution $p \in {\mathcal{P}}$, the space of all possible
distributions over $\mathcal{X}$. Perhaps the most popular choice
for a distance measure between distributions is the
Kullback-Leibler (KL) distance\endnote{Despite its popularity, the
KL distance $D(q\|p)$ between two probability distributions $q$
and $p$ is not a proper metric. It is not even symmetric between
$q$ and $p$. However it is non-negative, and equals zero only when
$q=p$.}: $D(q\|p) \triangleq \sum_{\mathbf{x}} q(\mathbf{x})\ln
\bigl(q(\mathbf{x}) / p(\mathbf{x})\bigr)$ \cite{coth91}. As the
KL distance is not symmetric in its arguments $p$ and $q$ we shall
refer to $D(q\|p)$ as the $qp$ KL distance (this is also sometimes
called the exclusive KL distance), and $D(p\|q)$ as the $pq$
distance (also sometimes called the inclusive KL distance).

Typically in physics $p$ is given by one of the statistical
``ensembles''. An important example of such KL minimization arises
with the Boltzmann distribution of the canonical ensemble:
$p(\mathbf{x})\propto\exp[-H(\mathbf{x})/T]$, where $H$ is the
``Hamiltonian'' of the system. The KL distance $D(q || p)$ to the
Boltzmann distribution is proportional to the Gibbs free energy of
statistical physics.  This free energy is identical to the maxent
Lagrangian considered in this paper. Stated differently, if one
solves for the distribution $q$ from one's set that minimizes $qp$
KL distance to the Boltzmann distribution, one gets the
distribution from one's set having maximal entropy, subject to the
constraint of having a specified expected value of $H$.  When the
set of distributions one's considering is $\mathcal{Q}$, the set
of product distributions, this $q$ minimizing $qp$ KL distance to
$p$ is called a ``mean-field approximation'' to $p$. So mean-field
theory is an instance of PC.

This illustrates that bounded rational games and the mean-field
approximation to Boltzmann distributions are essentially
identical. To relate them one equates $H$ with a common payoff
function $G$. The equivalence is completed by then identifying
each (independent) agent with a different one of the (independent)
physical variables in the argument of the
Hamiltonian.\endnote{Here and throughout, we fix the convention
that it is desirable to minimize objective functions, not to
maximize them.}

This connection between these fields allows us to exploit
techniques from statistical physics in bounded rational game
theory. For example, as mentioned above, rather than the canonical
ensemble, we can apply the grand canonical ensemble to bounded
rational games. This allows us to consider games in which the
number of players of each type is stochastic~\cite{wolp04a}.

\subsection{The contribution of this paper}

The use of a product distribution space $\mathcal{Q}$ for
optimization is consistent with game theory (and more generally
multi-agent systems). Further, this choice results in a highly
parallel algorithm, and is well-suited to problems that are
inherently distributed. Nonetheless, other concerns may dictate
different $\mathcal{Q}$.  In particular, in many optimization
tasks we seek multiple solutions which may be far apart from one
another. For example, in Constraint Satisfaction Problems (CSPs)
\cite{dechter03}, the goal is to identify all feasible solutions
which satisfy a set of constraints, or to show that none exist.
For small problem instances exhaustive enumeration techniques like
branch-and-bound are typically used to identify all such feasible
solutions. However, for larger problems it is desirable to develop
local-search-based approaches which determine multiple distinct
solutions in a single run.

In cases like these, where we desire multiple distinct solutions,
the use of PC with a product distribution is a poor choice. The
problem is that if each distribution $q_i$ is peaked about every
value of $x_i$ which occurs in at least one of the multiple
solutions, then in general there will be spurious peaks in the
product $q({\mathbf{x}}) = \prod q_i(x_i)$, i.e.,
$q({\mathbf{x}})$ may be peaked about some ${\mathbf{x}}$ that are
not solutions.  Alternatively, if each $q_i$ is peaked only about
a few of the solutions, this does not provide us with many
solutions. To address this we might descend the Lagrangian many
times, beginning from different starting points (i.e., different
initial $q$). However there is no guarantee that multiple runs
will each generate different solutions.

PC offers a simple solution to this problem that allows one to still
use product distributions: extend the event space underlying the
product distribution so that a single game provides multiple distinct
solutions to the optimization problem.  Intuitively speaking, such a
transformation recasts the problem in terms of a ``meta-game'' by
cloning the original game into several simultaneous games, with an
independent set of agents for each game. A supervisory agent chooses
what game is to be played.  We then form a Lagrangian for the
meta-game that is biased towards having any agents that control the
same variable in different games have different mixed strategies from
one another. The joint strategies for each of the separate games in
the meta-game then give a set of multiple solutions to the original
game. The supervisory agent sets the relative importance of which such
solution is used.  Since in general the resultant distribution across
the variables being optimized (i.e., across $\mathcal{Z}$) cannot be
written as a single product distribution, it provides coupling among
those variables.

Formally, the above process can be represented as a semicoordinate
transformation. Recall that the space of arguments to the
objective function $G(\mathbf{z})$ is $\mathcal{Z}$, and that the
product distribution is defined over $\mathcal{X}$. A
``semicoordinate system'' maps from $\mathcal{X}$ to
$\mathcal{Z}$~\cite{wobi04b,wolp04c}. Before introduction of the
semicoordinate system $\cal{X} = \cal{Z}$, and product
distributions over $\cal{X}$ give product distributions over
$\cal{Z}$. However when $\cal{X} \not= \mathcal{Z}$ and we
introduce a semicoordinate system, product distributions over
$\mathcal{X}$ (i.e., the noncooperative game is played in
$\mathcal{X}$) need not be product distributions on $\mathcal{Z}$.
By appropriate choice of the semicoordinate transformation, such
distributions can be made to correspond to any coupled
distributions across $\mathcal{Z}$.  In general, any Bayes net
topology can be achieved with an appropriate semicoordinate
transformation~\cite{wolp04c,wobi04b}. Different product
distributions over $\mathcal{X}$ correspond to different Bayes
nets having the same independence relations.

Here we consider a $\mathcal{X}$ that results in a mixture of $M$
product distributions $\mathcal{Z}$,
\begin{equation*}
q(\mathbf{z}) = \sum_{m=1}^M q^0(m) q^m(\mathbf{z}).
\end{equation*}
Intuitively, $q^0$ is the distribution over the moves of the
supervisor agent, with $m$ labelling the game that agent chooses.
This mixture of product distributions allows for the determination
of $M$ solutions at once. At the same time, an entropy term in the
Lagrangian ``pushes'' the separate products $q^m(\mathbf{z})$ in
the mixture apart. This biases the algorithm to locating well
separated solutions, as desired.

In Sec.~\ref{prodLDef} we review how one arrives at the Lagrangian
considered in this paper, the maxent Lagrangian. In
Sec.~\ref{minProdL} we review two elementary techniques introduced
in ~\cite{wobi04a,wolp03b,wolp04g} for updating a product
distribution $q$ to minimize the associated Lagrangian. Depending
on the form of the objective, the terms involved in the updating
of $q$ may be evaluated in closed form, or may require estimation
via Monte Carlo methods. In the experiments reported here all
terms are calculated in closed form. However, to demonstrate the
wider applicability of the update rules we review in
Appendix~\ref{sec:mc} a set of Monte-Carlo techniques providing low
variance estimates of required quantities. The first derivation of
these estimators is presented in this work.

With this background review complete, semicoordinate
transformations are introduced in Sec.~\ref{mixLDef}. As an
illustration of the use of semicoordinate transformations
particular attention is placed on mixture models, and how mixture
models may be seen as a product distributions over a different
space. In Sec.\ref{minimizeSect} we analyze the minimization of
the maxent Lagrangian associated with mixture-inducing
semicoordinate transformations. In that section we also relate our
maxent Lagrangian for mixture distributions to the Jensen-Shannon
distance over $\mathcal{X}$. Experimental validation of these
techniques is then presented for the $k$-satisfiability CSP
problem (section \ref{ksat}) and the $NK$ family of discrete
optimization problems (section \ref{nk}). These sections consider
the situation where the semicoordinate transformation is fixed a
priori, but suggestions are made on how to determine a good
semicoordinate transformation dynamically as the algorithm
progresses. We conclude with a synopsis of some other techniques
for updating a product distribution $q$ to minimize the associated
Lagrangian. This synopsis serves as the basis for a discussion of
the relationship between PC and other techniques.

Like all of PC, the techniques presented in this paper can readily
be applied to problems other than constrained optimization.  For
example, PC provides a natural improvement to the Metropolis
sampling algorithm \cite{wole04}, which the techniques of this
paper should be able to improve further. In addition, while for
simplicity we focus here on optimization over countable domains,
PC can be extended in many ways to continuous space optimization.
The associated technical difficulties can all be
addressed\cite{wost06}. See \cite{anbi04,wobi04b,biwo04b,biwo04c}
for other examples of PC and experiments.

It should be emphasized that PC usually is not a good choice for
how best to optimize problems lying in some narrowly defined
class. When one know a lot about the class of optimization
problems under scrutiny, algorithms that are hand-tailored for
that class will almost be called for. It is also in such
situations that one often can call upon formal convergence bounds.
In contrast, PC is in the spirit of Genetic Algorithms, the cross
entropy method, simulated annealing, etc. It is a broadly
applicable optimization algorithm that performs well in many
domains, even when there is little prior knowledge about the
domain. (See~\cite{woma97} for a general discussion of this
issue.)

Finally, in statistical inference, one parameterizes the possible
solutions to one's problem to reduce the dimensionality of the
solution space. Without such parameterization, the curse of
dimensionality prevents good performance, in general~\cite{duha00}.
By choosing one's parameterization though, one assumes (implicitly or
otherwise) that that parameterization is flexible enough to capture
the salient aspects of the stochastic process generating one's
data. In essence, one assumes that the parameterization ``projects
out'' the noise while keeping the signal.

The PC analogue of the problem of what parameterization to use is what
precise semicoordinate transformation to use. Just as there is no
universally correct choice of how to parameterize a statistics
problem, there is no universally correct choice of what semicoordinate
transformation to use. In both situations, one must rely on prior
knowledge to make one's choice, potentially combined with conservative
online adaptation of that choice.

\section{The Lagrangian for Product Distributions} \label{prodLDef}

We begin by considering the case of the identity semicoordinate
system, and ${\mathcal{X}} = {\mathcal{Z}}$. As discussed above,
we consider $qp$ distance to the $T$-parameterized Boltzmann
distribution $p(\mathbf{x})=\exp[-G(\mathbf{x})/T] / Z(T)$ where
$Z(T)$ is the normalization constant. At low $T$ the Boltzmann
distribution is concentrated on $\mathbf{x}$ having low $G$
values, so that the product distribution with minimal $qp$
distance to it would be expected to have the same behavior.
Accordingly, one would expect that by taking $qp$ KL distance to
this distribution as one's Lagrangian, and modifying the
Lagrangian from one iteration to the next by lowering $T$, one
should end up at a $q$ concentrated on $\mathbf{x}$ having low $G$
values. (See \cite{wobi04a,wolp04g} for a more detailed formal
justification of using this Lagrangian based on solving
constrained optimization problems with Lagrange parameters.)

More precisely, the $qp$ KL distance to the Boltzmann distribution
is the maxent Lagrangian,
\begin{equation}
\mathcal{L}(q) = \mathbb{E}_q(G) - T S(q) \label{freeEnergyEq}
\end{equation}
up to irrelevant additive and multiplicative constants.
Equivalently, we can write it as
\begin{equation}
\mathcal{L}(q) = \beta \mathbb{E}_q(G) -  S(q) \label{altfreeEnergyEq}
\end{equation}
where $\beta \triangleq 1/T$, up to an irrelevant overall constant.
In these equations the inner product $\mathbb{E}_q(G) \triangleq
\sum_\mathbf{x} q(\mathbf{x}) G(\mathbf{x})$ is the expected value
of $G$ under $q$, and $S(q)\triangleq-\!\sum_{\mathbf{x}}
q(\mathbf{x}) \ln q(\mathbf{x})$ is the Shannon entropy of $q$.

For $q$'s which are product distributions $S(q)=\sum_i S(q_i)$
where $S(q_i) = -\!\sum_{x_i} q_i(x_i) \ln q_i(x_i)$. Accordingly,
we can view the maxent Lagrangian as equivalent to a set of
Lagrangians, $\mathcal{L}_i(q) = \sum_{x_i}
\mathbb{E}_{q_{-i}}[G(x_i, \mathbf{x}_{-i})] q_i(x_i) - T
S_i(q_i)$, one such Lagrangian for each agent $i$ so that
$\mathcal{L}(q) = \sum_{i=1}^n \mathcal{L}_i(q)$.\endnote{We adopt
the notation that $q_{-i}$ indicates the distribution $q$ with
variable $i$ marginalized out, i.e., the product $\prod_{j \ne i}
q_j$. Analogously, $\mathbf{x}_{-i} \triangleq [x_1, \cdots,
x_{i-1}, x_{i+1}, \cdots, x_n]$} The first term in $\mathcal{L}$
is minimized by having perfectly rational players, i.e. by players
who concentrate all their probability on the moves that are best
for them, given the distributions over the agents. The second term
is minimized by perfectly irrational players, i.e., by a perfectly
uniform joint mixed strategy $q$. So $T$ specifies the balance
between the rational and irrational behavior of the players. In
particular, for $T \rightarrow 0$, by minimizing the Lagrangian we
recover the Nash equilibria of the game. Alternatively, from a
statistical physics perspective, where $T$ is the temperature of
the system, this maxent Lagrangian is simply the Gibbs free energy
for the Hamiltonian $G$.

\subsection{Incorporating constraints}

Since we are interested in problems with constraints, we replace
$G$ in Eqs.~\eqref{freeEnergyEq} and \eqref{altfreeEnergyEq} with
\begin{equation}
G(\mathbf{x}) + \sum_{a=1}^C \lambda_a
c_a(\mathbf{x})
\label{eq:constraintG}
\end{equation}
where $G$ is the original objective function and the $c_a$ are the
set of $C$ equality constraint functions that are required to be
equal to zero. For constraint satisfaction problems we take the
original objective function to be the constant function 0. The
$\lambda_a$ are Lagrange multipliers that are used to enforce the
constraints. Collectively, we refer to the Lagrange multipliers
with the $C\times 1$ vector $\pmb{\lambda}$. The constraints are
all equality constraints, so a saddle point of the Lagrangian over
the space of possible $q$ and $\pmb{\lambda}$ is a solution of our
problem. Note however that we do not have to find the exact saddle
point; in general sampling from a $q$ close to the saddle point
will give us the ${\mathbf{x}}$'s we seek.

There are certainly other ways in which constraints can be
addressed within the PC framework. An alternative approach might
allow constraints to be weakly violated. We would then iteratively
anneal down those weaknesses, i.e., strengthen the constraints, to
where they are not violated. In this approach we could replace the
maxent Lagrangian formulation encapsulated in Eq.'s
~\eqref{altfreeEnergyEq} and ~\eqref{eq:constraintG} with
\begin{equation}
\mathcal{L}(q, \beta, \pmb{\lambda}) = \beta [\mathbb{E}_q(G) -
\gamma_G] + \sum_a \lambda_a [\mathbb{E}_q(c_a) - \gamma_a] - S(q) .
\end{equation}
In each iteration of the algorithm  $\beta$, $\pmb{\lambda}$ are
treated as Lagrange parameters and one solves for their values that
enforce the equality constraints $\mathbb{E}_q(G) = \gamma_G$, and
the $C$ constraints $\mathbb{E}_q(c_a) = \gamma_a$ while also
minimizing $\mathcal{L}(q, \beta, \pmb{\lambda})$. In the usual way,
since our constraints are all equalities, one can do this by finding
saddle points of $\mathcal{L}(q, \beta, \pmb{\lambda})$. The next
iteration would then start by modifying our Lagrangian by shrinking
the values $\gamma_G$, \{$\gamma_a$\} slightly before proceeding to
a new process of finding a saddle point.

Another more theoretically justified way to incorporate
constraints requires that the support of $q$ is constrained to lie
entirely within the feasible region. Any $\mathbf{x}$ which
violates a constraint is assigned 0 probability, i.e.
$q(\mathbf{x})=0$ at all $\mathbf{x}$ which violate the
constraints.

For pedagogical simplicity, we do not consider these alternative
approaches, but concentrate on the Lagrangian of
Eq.~\eqref{freeEnergyEq} with the $G$ of
Eq.~\eqref{eq:constraintG}. In addition to the constraints
associated with the optimization problem the vectors \{$q_i$\}
must be probability distributions. So there are implicit
constraints our solution must satisfy: $0 \le q_i(x_i)\le 1$ for
all $i$ and $x_i$, and $\sum_{x_i} q_i(x_i) = 1$ for all $i$. To
reduce the size of our equations we do not explicitly write these
constraints.

\section{Minimizing the maxent Lagrangian}\label{minProdL}

For fixed $\beta$, our task is to find a saddle point of
${\mathcal{L}}(q, \pmb{\lambda})$.  In ``first order methods'' a
saddle point is found by iterating a two-step process. In the
first step the Lagrange parameters $\pmb{\lambda}$ are fixed and
one solves for the $q$ that minimizes the associated
$\mathcal{L}$.{\endnote{Properly speaking one should find the
global minimizer $q$. Here we content ourselves with finding local
minima.}}  In the second step one then freezes that $q$ and
updates the Lagrange parameters. There are more sophisticated ways
of finding saddle points \cite{gran05}, and more generally one can
use modified versions of the Lagrangian (e.g., an augmented
Lagrangian \cite{bert96}). For simplicity we do not consider such
more sophisticated approaches.

In this section we review two approaches to finding the \{$q_i$\}
for fixed Lagrange multipliers $\pmb{\lambda}$. We also describe our
approach for the second step of the first order method, i.e., we
describe how we use gradient ascent to update the Lagrange
multipliers $\lambda_a$ for fixed $q$. See
\cite{wolp03b,wobi04a,wolp04g} for further discussion of these
approaches as well as the many others one can use.

\subsection{Brouwer Updating}

At each step $t$ the direction in the simplex $\mathcal{Q}$ that, to
first order, maximizes the drop in $\mathcal{L}$ is given by (-1
times)
\begin{equation}
{\tilde{\pmb{\nabla}}}_q{\mathcal{L}}(q) \triangleq
\pmb{\nabla}_q{\mathcal{L}}(q) - \pmb{\eta}(q). \label{eq:nabla}
\end{equation}
In this equation $\pmb{\eta}(q)$ is proportional to the unit vector,
with its magnitude set to ensure that a step in the direction
${\tilde{\pmb{\nabla}}}_q{\mathcal{L}}(q)$ remains in the unit
simplex.  Furthermore, the $q_i(x_i)$ component of the gradient,
one for every agent $i$ and every possible move $x_i$ by the agent,
is (up to constant terms which have been absorbed into
$\pmb{\eta}(q)$):
\begin{equation}
[\nabla_q{\mathcal{L}}(q)]_{q_i(x_i)} = \frac{\partial
\mathcal{L}}{\partial q_i(x_i)} = \mathbb{E}_{q_{- i}}\!(G|x_i) + T
\ln[q_i(x_i)] \label{gradLEq}
\end{equation}
where
\begin{equation}
\mathbb{E}_{q_{- i}} G|x_i) = \sum_{\mathbf{x}_{- i}} q_{-
i}(\mathbf{x}_{- i}) G(x_i,\mathbf{x}_{- i}) \label{condExpEq}
\end{equation}
with $\mathbf{x}_{-i} \triangleq [x_1, \cdots, x_{i-1}, x_{i+1},
\cdots, x_n]$ and $q_{- i}(\mathbf{x}_{- i}) \triangleq
\prod_{j=1|j\not=i}^n q_j(x_j)$. $\pmb{\eta}(q)$ is the vector
that needs to be added to $\pmb{\nabla}_q \mathcal{L}(q)$ so that
each $q_i$ is properly normalized.{\endnote{N.b., we do \emph{not}
project onto $\mathcal{Q}$ but rather add a vector to get back to
it. See ~\cite{wobi04a}.}} The $q_i(x_i)$ component of
$\pmb{\eta}(q)$, is equal to
\begin{equation}
[\eta(q)]_{q_i(x_i)} = \eta_i(q) \triangleq
\frac{1}{|{\mathcal{X}}_i|} \sum_{x'_i} \; [\nabla_q
\mathcal{L}(q)]_{q_i(x'_i)} \label{etaEq}
\end{equation}
where $|{\mathcal{X}}_i|$ is the number of possible moves (allowed
values) $x_i$. Not that for any agent $i$, all of the associated
components of $\pmb{\eta}(q)$, namely $q_i(x_1), \cdots,
q_i(x_{|\mathcal{X}_i|})$, share the same value $\eta_i(q)$. This
choice ensures that $\sum_{x_i} q_i(x_i)=1$ after the gradient
update to the values $q_i(x_i)$.

The expression in Eq.~\eqref{condExpEq} is the expected payoff to
agent $i$ when it plays move $x_i$, under the distribution $q_{-i}$
across the moves of all other agents. Setting
$\tilde{\nabla}_q{\mathcal{L}}(q)$ to zero gives the solution
\begin{equation}
q_i^{t+1}(x_i) \propto \exp\bigl[-\mathbb{E}_{q^t_{-
i}}\!(G|x_i)/T\bigr]
\label{BrouwerEq}
\end{equation}
Brouwer's fixed point theorem guarantees the solution of
Eq.~\eqref{BrouwerEq} exists for any $G$~\cite{wolp04a,wolp03b}.
Hence we call update rules based on this equation {\emph{Brouwer
updating}}.

Brouwer updating can be done in parallel on all the agents.
However, one problem that can arise if all agents update in
parallel is ``thrashing''. In Eq. \eqref{BrouwerEq} each agent $i$
adopts the $q_i$ that is be optimal assuming the other agents
don't change their distributions. However, other agents do change
their distributions, and thereby at least partially confound agent
$i$. One way to address this problem is to have agent $i$ not use
the current value $\mathbb{E}_{q^t_{- i}}\!(G|x_i)$ alone to
update $q^t_i(x_i)$, but rather use a weighted average of all
values $\mathbb{E}_{q^{t'}_{- i}}\!(G|x_i)$ for $t' \le t$, with
the weights shrinking the further into the past one goes. This
introduces an inertia effect which helps to stabilize the
updating. (Indeed, in the continuum-time limit, this weighting
becomes the replicator dynamics \cite{wolp04c}.)

A similar idea is to have agent $i$ use the current
$\mathbb{E}_{q^t_{- i}}\!(G|x_i)$ alone, but have it only move part of
the way the parallel Brouwer update recommends. Whether one moves all
the way or only part-way, what agent $i$ is interested in is what
distribution will be optimal {\it{for the next distributions of the
other agents}}. Accordingly, it makes sense to have agent $i$ predict,
using standard time-series tools, what those future distributions will
be. This amounts to predicting what the next vector of values of
$\mathbb{E}_{q^t_{- i}}\!(G|x_i)$ will be, based on seeing how that
vector has evolved in the recent past. See \cite{shar04} for related ideas.

Another way of circumventing thrashing is to have the agents
update their distributions serially (one after the other) rather
than in parallel. See~\cite{wobi04b} for a description of various
kinds of serial schemes, as well as a discussion of partial
serial, partial parallel algorithms.

\subsection{Nearest-Newton Updating}

To evaluate the gradient one only needs to evaluate or estimate
the terms $\mathbb{E}_{q^t_{-i}}(G | x_i)$ for all agents (see
below and ~\cite{wolp03b,wolp04a}). Consequently, gradient descent
is typically straight-forward. Though, it is also usually simple
to evaluate the Hessian of the Lagrangian, conventional Newton's
descent is intractable for large systems because inverting the
Hessian is computationally expensive.

Of course there are schemes such as conjugate gradient or
quasi-Newton that do exploit second order information even when
the Hessian cannot be inverted.  However, the special structure of
the Lagrangian also allows second order information to be used for
a simple variant of Newton descent. The associated update rule is
called \emph{Nearest-Newton} updating~\cite{wobi04a}; we review it
here.

To derive Nearest-Newton we begin by considering the Lagrangian
$\mathbb{E}_\pi(G) - TS(\pi)$, for an $unrestricted$ probability
distribution $\pi$.\endnote{For such a distribution we relax the
requirement of being a product or having any other particular
form; we only require that all $0\le \pi(\mathbf{x})\le 1$ and
$\sum_{\mathbf{x}} \pi(\mathbf{x})=1$.} This Lagrangian is a
convex function of $\pi$ with a diagonal Hessian. So given a
current distribution $\pi^t$ we can make an unrestricted Newton
step of this Lagrangian to a new distribution $\pi^{t+1}$. That
new distribution typically is not in $\mathcal{Q}$ (i.e. not a
product distribution), even if the starting distribution is.
However we can solve for the $q^{t+1}\in\mathcal{Q}$ that is
nearest to $\pi^{t+1}$, for example by finding the
$q^{t+1}\in\mathcal{Q}$ that minimizes $qp$ KL distance $D(p||q)$
to that new point.

More precisely, the Hessian of $\mathbb{E}_\pi(G) - TS(\pi)$,
$\partial^2 \mathcal{L} / \partial \pi(\mathbf{x})
\partial \pi(\mathbf{x}')$, is diagonal, and so is simply
inverted. This gives the Newton update for $\pi^t$:
\begin{equation*}
\pi^{t+1}(\mathbf{x}) = \pi^t(\mathbf{x}) - \alpha^t_q
\pi^t(\mathbf{x})\left[ \frac{G(\mathbf{x}) -
\mathbb{E}_{\pi^t}\!(G)}{T} + S(\pi^t) + \ln \pi^t(\mathbf{x})
\right]
\end{equation*}
which is normalized if $\pi^t$ is normalized and where $\alpha^t_q$
is a step size. As $\pi^t$ will typically not belong to
$\mathcal{Q}$ we find the product distribution nearest to
$\pi^{t+1}$ by minimizing the KL distance $D(\pi^{t+1}\|q)$ with
respect to $q$. The result is that $q_i(x_i) = \pi^{t+1}_i(x_i)$,
i.e. $q_i$ is the marginal of $\pi^{t+1}$ given by integrating it
over $\mathbf{x}_{-i}$.

Thus, whenever $\pi^t$ itself is a product distribution, the update
rule for $q_i(x_i)$ is
\begin{equation}
q_i^{t+1}(x_i) = q_i^t(x_i) - \alpha^t_q q_i^t(x_i) \biggl[
\frac{\mathbb{E}_{q^t_{- i}}\!(G|x_i) - \mathbb{E}_{q^t}\!(G)}{T} +
S(q_i) + \ln q_i^t(x_i) \biggr]. \label{nearNewtEq}
\end{equation}
This update maintains the normalization of $q_i$, but may make one
or more $q_i^{t+1}(x_i)$ greater than 1 or less than 0. In such
cases we set $q^{t+1}_i$ to be valid product distribution nearest in
Euclidean distance (rather than KL distance) to the suggested Newton
update.

\subsection{Updating Lagrange Multipliers}

In order to satisfy the imposed optimization constraints
$\{c_a(\mathbf{x})\}$ we must also update the Lagrange multipliers.
To minimize communication between agents this is done in the
simplest possible way -- by gradient descent. Taking the partial
derivatives with respect to $\lambda_a$ gives the update rule
\begin{equation}
\lambda_a^{t+1} = \lambda_a^t + \alpha_\lambda^t
\mathbb{E}_{q^t_*}\bigl(c_a(\mathbf{x})\bigr)
\label{eq:updateconstrain}
\end{equation}
where $\alpha^t_\lambda$ is a step size and $q^t_*$ is the local
minimizer of $\mathcal{L}$ determined as above at the old settings,
$\pmb{\lambda}^t$, of the multipliers.

\subsection{Other descent schemes}

It should be emphasized that PC encompasses many approaches to
optimization of the Lagrangian that differ from those used here.  For
example, in \cite{biwo04a,wolp04c} there is discussion of alternative
types of descent algorithms that are related to block relaxation, as
well as to the fictitious play algorithm of game theory
\cite{futi91,shar04} and multi-agent reinforcement learning algorithms
like those in collective intelligence \cite{wotu01a,wotu03a}.

As another example, see \cite{wobi04a,wolp04a} for discussions of
using $pq$ KL distance (i.e., $D(p || q)$) rather than $qp$
distance. Interestingly, as discussed below, that alternative
distance must be used even for descent of $qp$ distance, if one
wishes to use 2nd order descent schemes.  \cite{wolp04g} discusses
using non-Boltzmann target distributions $p$, and many other
options for what functional(s) to descend.

\subsection{Algorithmic summary}

Having described two possible PC algorithms we summarize the steps
involved in each. This basic framework will form the basis for the
semicoordinate extensions described in Section \ref{mixLDef}.
\begin{algorithm}
\dontprintsemicolon \SetKwData{updateQ}{updateQ}
\SetKwData{updateLambda}{update$\lambda$}
\SetKwData{updateT}{updateT}
\SetKwData{initMult}{initializeMultipliers}
\SetKwData{initQ}{initializeQ} \SetKwData{initT}{initializeT}
\SetKwData{constraintsSat}{constraintsSatisfied}
\SetKwData{evalExp}{evalConditionalExpectations}
\SetKwData{atLocalMin}{atLocalMinimum($q$)} \SetKwData{done}{done}

\KwIn{An objective function $G(\mathbf{x})$, and a set of equality
constraint functions $\{c_a(\mathbf{x})=0\}_{a=1}^C$} \KwOut{A
product distribution $q^*(\mathbf{x})=\prod_i q_i^*(x_i)$ peaked
about the minimizer of $G$ satisfying the constraints.}

\BlankLine $\pmb{\lambda} \leftarrow$ \initMult \; $T \leftarrow$
\initT \; $q \leftarrow$ \initQ \;

\Repeat{\done}{ \Repeat{\constraintsSat}{ \Repeat{\atLocalMin}{$gs
\leftarrow$ \evalExp($G, q$) \; $q \leftarrow$ \updateQ($gs$) }
$\pmb{\lambda} \leftarrow$ \updateLambda } $T \leftarrow$ \updateT
} \caption{The basic PC algorithmic framework.\label{basicAlg}}
\end{algorithm}
Pseudocode for basic PC optimization appears in Algorithm
\ref{basicAlg}. Lines 1 -- 3 initialize algorithmic parameters.
The best starting temperatures and multiplier values vary from
problem to problem. We typically initialize $q$ to be the maximum
entropy distribution which is uniform over the search space
$\mathcal{X}$. An outer loop decreases the temperature according
to a a schedule determined by the function updateT. Later we
comment on automatic schedules generated from the settings of
Lagrange multipliers. Inside this loop is another loop which
increments Lagrange multipliers according to Eq.
\eqref{eq:updateconstrain} every time $q$ is iterated to a local
minimum of the Lagrangian. The minimization of $L$ for a fixed
temperature and setting of multipliers is accomplished in the
innermost loop. This minimization is accomplished by repeatedly
determining all the conditional probabilities $\mathbb{E}_{q_{-
i}}(G+\sum_a \lambda_a c_a|x_i)$ (line 7), and then using these in
either of the two update rules Eqs. \eqref{BrouwerEq},
\eqref{nearNewtEq} (line 8). The evaluation of the conditional
expectations of $G + \sum_a \lambda_a c_a$ can be accomplished
either analytically or with Monte Carlo estimates. For many
problems analytical evaluation is prohibitively costly, and Monte
Carlo methods are the only option. In Appendix A we consider
unbiased low variance Monte Carlo methods for estimating the
required conditional expectations, and in Appendix B we derive the
minimal variance estimator from within a class of useful
estimators.

\section{Semicoordinate Transformations}
\label{mixLDef}

\subsection{Motivation}

Consider a multi-stage game like chess, with the stages (i.e., the
instants at which one of the players makes a move) delineated by
$t$. In game theoretic terms,  the ``strategy'' of a player is the
mapping from board-configuration to response that specifies the rule
it adopts before play starts
\cite{futi91,baol99,osru94,auha92,fule98}. More generally, in a
multi-stage game like chess the strategy of player $i$, $x_i$, is
the set of $t$-indexed maps taking what that player has observed in
the stages $t' < t$ into its move at stage $t$. Formally, this set
of maps is called player $i$'s {\bf{normal form}} strategy.

The joint strategy of the two players in chess sets their joint
move-sequence, though in general the reverse need not be true. In
addition, one can always find a joint strategy to result in any
particular joint move-sequence.  Now typically at any stage there is
overlap in what the players have observed over the preceding stages.
This means that even if the players' strategies are statistically
independent (being separately set before play started), their move
sequences are statistically coupled.  In such a situation, by
parameterizing the space $\mathcal{Z}$ of joint-move-sequences
$\mathbf{z}$ with joint-strategies $\mathbf{x}$, we shift our focus
from the coupled distribution $P(\mathbf{z})$ to the decoupled
product distribution, $q(\mathbf{x})$. This is the advantage of
casting multi-stage games in terms of normal form strategies.

More generally, given any two spaces $\mathcal{X}$ and
$\mathcal{Z}$, any associated onto mapping $\zeta: \mathcal{Z}
\rightarrow \mathcal{X}$, not necessarily invertible, is called a
{\it{semicoordinate system}}. The identity mapping $id:
\mathcal{Z} \rightarrow \mathcal{Z}$ is a trivial example of a
semicoordinate system. Another semicoordinate system is the
mapping from joint-strategies in a multi-stage game to joint
move-sequences. In other words, changing the representation space
of a multi-stage game from move-sequences $\mathbf{z}$ to
strategies $\mathbf{x}$ is a semicoordinate transformation of that
game.

Intuitively, a semicoordinate transformation is a
reparameterization of how a game --- a mapping from joint moves to
associated payoffs --- is represented.  So we can perform a
semicoordinate transformation even in a single-stage game.  Say we
restrict attention to distributions over $\mathcal{X}$ that are
product distributions.  Then changing $\zeta(\cdot)$ from the
identity map to some other function means that the players' moves
are no longer independent. After the transformation their move
choices --- the components of $z$ --- are statistically coupled,
even though we are considering a product distribution.

Formally, this is expressed via the standard rule for transforming
probabilities,
\begin{eqnarray}
P_{\mathcal{Z}}(\mathbf{z} \in \mathcal{Z}) \triangleq
\zeta(P_{\mathcal{X}}) \triangleq \int d\mathbf{x} \;
P_{\mathcal{X}}(\mathbf{x}) \delta(\mathbf{z} -
\zeta(\mathbf{x})), \label{eq:semitransf}
\end{eqnarray}
\noindent where $P_{\mathcal{X}}$ and $P_{\mathcal{Z}}$ are the
distributions across $\mathcal{X}$ and $\mathcal{Z}$,
respectively. To see what this rule means geometrically, recall
that $\mathcal{P}$ is the space of all distributions (product or
otherwise) over $\mathcal{Z}$ and that $\mathcal{Q}$ is the space
of all product distributions over $\mathcal{X}$. Let
$\zeta({\mathcal{Q}})$ be the image of $\mathcal{Q}$ in
$\mathcal{P}$. Then by changing $\zeta(\cdot)$, we change that
image; different choices of $\zeta(\cdot)$ will result in
different manifolds $\zeta({\mathcal{Q}})$.

As an example, say we have two players, with two possible moves
each. So $\mathbf{z}$ consists of the possible joint moves,
labelled $(0, 0), (0, 1), (1, 0)$ and $(1, 1)$. Take $\mathcal{X}
= \mathcal{Z}$, and choose $\zeta(0, 0) = (0, 0), \; \zeta(0, 1) =
(1, 1), \; \zeta(1, 0) = (1, 0)$, and $\zeta(1, 1) = (0, 1)$. Say
that $q$ is given by $q_1(x_1 = 0) = q_2(x_2 = 0) = 2/3$. Then the
distribution over joint-moves $\mathbf{z}$ is $P_{\mathcal{Z}}(0,
0) = P_{\mathcal{X}}(0, 0) = 4/9$, $P_{\mathcal{Z}}(1, 0) =
P_{\mathcal{Z}}(1, 1) = 2/9$, $P_{\mathcal{Z}}(0, 1) = 1/9$. So
$P_{\mathcal{Z}}(\mathbf{z}) \ne
P_{\mathcal{Z}}(z_1)P_{\mathcal{Z}}(z_2)$; the moves of the
players are statistically coupled, even though their strategies
$x_i$ are independent.

Any $P_{\mathcal{Z}}$, no matter what the coupling among its
components, can be expressed as $\zeta(P_{\mathcal{X}})$ for some
product distribution $P_{\mathcal{X}}$ for and associated
$\zeta(\cdot)$ In the worst case, one can simply choose
$\mathcal{X}$ to have a single component, with $\zeta(\cdot)$ a
bijection between that component and the vector $z$ --- trivially,
any distribution over such an $\mathcal{X}$ is a product
distribution. Another simple example is where one aggregates one
or more agents into a new single agent, i.e., replaces the product
distribution over the joint moves of those agents with an
arbitrary distribution over their joint moves. This is related to
the concept coalitions in cooperative game theory, as well as to
Aumann's correlated equilibrium \cite{futi91,auma87,auha92}.

Less trivially, given any model class of distributions
\{$P_{\mathcal{Z}}$\}, there is an $\mathcal{X}$ and associated
$\zeta(\cdot)$ such that \{$P_{\mathcal{Z}}$\} is identical to
$\zeta({\cal{Q}}_X)$. Formally this is expressed in a result
concerning Bayes nets. For simplicity, restrict attention to
finite $\mathcal{Z}$. Order the components of $\mathcal{Z}$ from 1
to $N$. For each index $i \in \{1, 2, \ldots, N \}$, have the
parent function $\Pa(i, \mathbf{z})$ fix a subset of the
components of $\mathbf{z}$ with index greater than $i$, returning
the value of those components for the $\mathbf{z}$ in its second
argument if that subset of components is non-empty. So for
example, with $N > 5$, we could have $\Pa(1, \mathbf{z}) = (z_2,
z_5)$. Another possibility is that $\Pa(1, \mathbf{z})$ is the
empty set, independent of $\mathbf{z}$.

Let $A(\Pa)$ be the set of all probability distributions
$P_{\mathcal{Z}}$ that obey the conditional independencies implied
by $\Pa$: $\forall \; P_{\mathcal{Z}} \in A(\Pa), \mathbf{z} \in
\mathcal{Z},$
\begin{eqnarray}
P_{\mathcal{Z}}(\mathbf{z}) &=& \prod_{i=1}^N P_{\mathcal{Z}}(z_i
| \Pa(i, \mathbf{z})). \label{eq:bayesnet}
\end{eqnarray}
By definition, if $\Pa(i, \mathbf{z}))$ is empty, $
P_{\mathcal{Z}}(z_i | \Pa(i, \mathbf{z}))$ is just the $i$'th
marginal of $P_{\mathcal{Z}}$, $P_{\mathcal{Z}}(z_i)$. As an
example of these definitions, the dependencies
$\{\Pa(1,\mathbf{z}) = (z_2,z_3), \Pa(2,\mathbf{z})=z_4,
\Pa(3,\mathbf{z})=(), \Pa(4,\mathbf{z})=() \}$ correspond to the
family of distributions factoring as
\begin{equation*}
P(\mathbf{z}) = P(z_1|z_2,z_3)P(z_2|z_4)P(z_3) P(z_4)
\end{equation*}

As proven in \cite{wobi04b}, for any choice of $\Pa$ there is an
associated set of distributions $\zeta({\cal{Q}}_X)$ that equals
$A(\Pa)$ exactly:

\noindent \textbf{Proposition:} Define the components of $X$ using
multiple indices: For all $i \in \{1, 2, \ldots, N\}$ and possible
associated values (as one varies over $\mathbf{z} \in \mathcal{Z}$)
of the vector $\Pa(i, \mathbf{z})$, there is a separate component of
$\mathbf{x}$, $x_{i; \Pa(i, \mathbf{z})}$. This component can take
on any of the values that $z_i$ can. Define $\zeta(\cdot)$
recursively, starting at $i = N$ and working to lower $i$, by the
following rule: $\forall \; i \in \{1, 2, \ldots, N\}$,
\begin{eqnarray*}
[\zeta(\mathbf{x})]_i &=& x_{i; \Pa(i, \mathbf{z})} .
\end{eqnarray*}
Then $A(\Pa) = \zeta({\cal{Q}}_X)$.

Intuitively, each component of $\mathbf{x}$ in the proposition is
the conditional distribution $ P_{\mathcal{Z}}(z_i | \Pa(i,
\mathbf{z}))$ for some particular instance of the vector $\Pa(i,
\mathbf{z})$. As illustration consider again the example
$\{\Pa(1,\mathbf{z}) = (z_2,z_3), \Pa(2,\mathbf{z})=z_4,
\Pa(3,\mathbf{z})=(), \Pa(4,\mathbf{z})=() \}$. If each $z_i$
assumes the value 0 or 1, then $\mathbf{x}$ has 8 components $x_4,
x_3, x_{2;0}, x_{2;1}, x_{1;00}, x_{1;01}, x_{1;10}$, and
$x_{1;11}$ with each component also either 0 or 1. The product
distribution in $\mathcal{X}$ is
\begin{equation*}
q(\mathbf{x}) = q_4(x_4) q_3(x_3) q_{2;0}(x_{2;0}) q_{2;1}(x_{2;1})
q_{1;00}(x_{1;00}) q_{1;01}(x_{1;01}) q_{1;10}(x_{1;10})
q_{1;11}(x_{1;11}).
\end{equation*}
Under $\zeta$ the distribution $q_4(x_4)$ is mapped to $q_4(z_4)$,
$q_{2;0}(x_{2;0})$ is mapped to $q_2(z_2|z_4=0)$,
$q_{1;01}(x_{1;01})$ is mapped to $q_1(z_1|z_2=0,z_3=1)$, and so on.

The proposition means that in principle we never need consider
coupled distributions. It suffices to restrict attention to
product distributions, so long as we use an appropriate
semicoordinate system. Semicoordinate systems also enable the
representation of mixture models over $\mathcal{Z}$ can be also be
represented using products. However, before discussing mixture
models we show how transformation of semicoordinate systems can in
principle be used to escape local minima in $L(q)$.

\subsection{Semicoordinate transformations and local minima}

To illustrate another application of semicoordinate transformations,
we confine ourselves to the case where $\mathcal{X} = \mathcal{Z}$
so that $\zeta$ is a bijection on $\mathcal{X}$.

We assume that the domain of the $i$th of $n$ variables has size
$|\mathcal{X}_i|$. Then $|\mathcal{X}| = \prod_{i=1}^n
|\mathcal{X}_i|$ is the size of the search space. Each coordinate
variable $x_i$ partitions the search space into $|\mathcal{X}_i|$
disjoint regions. The partitions are such that the intersection over
all variable coordinates yields a single ${\mathbf{x}}$. In
particular, the standard semicoordinate system relies on the
partition $[*, \, \cdots, \, *, x_i=0, *, \cdots, *]$, $\cdots$,
$[*, \, \cdots, \, *, x_i=|\mathcal{X}_i|-1, *, \cdots, *]$ for each
coordinate $\mathbf{x}_i$.

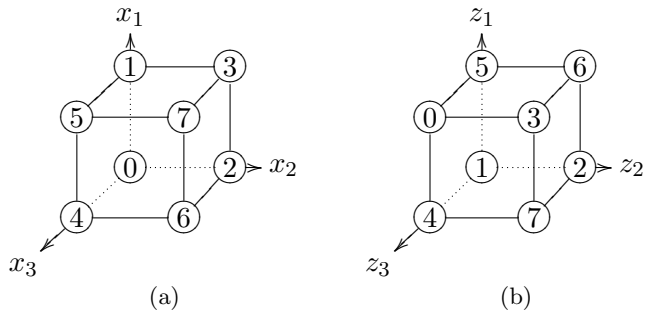
\begin{figure}
\begin{center}
\mbox{ \subfigure[]{\xymatrix@-1.5pc{ & & x_1 & & & & \\
& & *+[o][F-]{1} \ar@{-}[rr] \ar@{.}[dd] \ar[u]& & *+[o][F-]{3} \ar@{-}[dd] & \\
& *+[o][F-]{5} \ar@{-}[rr]\ar@{-}[ur]\ar@{-}[dd] & & *+[o][F-]{7} \ar@{-}[dd]\ar@{-}[ur] &  & \\
&  & *+[o][F-]{0} \ar@{.}[rr] & & *+[o][F-]{2} \ar[r] & x_2 \\ &
 *+[o][F-]{4} \ar[dl] \ar@{-}[rr] \ar@{.}[ur] & & *+[o][F-]{6} \ar@{-}[ur]& & \\
 x_3 & & & & &}}}
\mbox{ \subfigure[]{\xymatrix@-1.5pc{ & & z_1 & & & & \\
& & *+[o][F-]{5} \ar@{-}[rr] \ar@{.}[dd] \ar[u]& & *+[o][F-]{6} \ar@{-}[dd] & \\
& *+[o][F-]{0} \ar@{-}[rr]\ar@{-}[ur]\ar@{-}[dd] & & *+[o][F-]{3} \ar@{-}[dd]\ar@{-}[ur] &  & \\
& & *+[o][F-]{1} \ar@{.}[rr] & & *+[o][F-]{2} \ar[r] & z_2 \\
& *+[o][F-]{4} \ar[dl] \ar@{-}[rr] \ar@{.}[ur] & & *+[o][F-]{7} \ar@{-}[ur]& & \\
 z_3 & & & & &}}}
\end{center}
\caption{(a) Original linear indexing for 3 binary variables $x_1,
x_2, x_3$. (b) Result after applying the transformation to the new
variables $z_1, z_2, z_3$.} \label{coordFig}
\end{figure}

As a illustrative example, consider 3 binary variables where
$\mathcal{X}=\{0,1\}^3$. Figure \ref{coordFig}(a) shows the 8 points
in the search space represented in the standard coordinate system.
Figure \ref{coordFig}(b) shows a shuffling of the 8 configurations
under the permutation $(0 \, 1 \, 2 \, 3 \, 4 \, 5 \, 6 \, 7)
\overset{\zeta}{\rightarrow} (1 \, 5 \, 2 \, 6 \, 4 \, 0 \, 7 \,3)$.
The resulting partitions of configurations are given in Table
\ref{coordPPTable}.
\begin{table}
\begin{center}
\begin{tabular}{|c|c|c|c|}
\hline
$x_1=0$ & (0,0,0), (0,0,1), (0,1,0), (0,1,1) & $z_1=0$ & (1,0,0), (0,1,0), (0,0,1), (1,1,1) \\
$x_1=1$ & (1,0,0), (1,0,1), (1,1,0), (1,1,1) & $z_1=1$ & (0,0,0), (1,1,0), (1,0,1), (0,1,1) \\
\hline
$x_2=0$ & (0,0,0), (0,0,1), (1,0,0), (1,0,1) & $z_2=0$ & (1,0,0), (0,1,0), (1,0,1), (0,1,1) \\
$x_2=1$ & (0,1,0), (0,1,1), (1,1,0), (1,1,1) & $z_2=1$ & (0,0,0), (1,1,0), (0,0,1), (1,1,1) \\
\hline
$x_3=0$ & (0,0,0), (0,1,0), (1,0,0), (1,1,0) & $z_3=0$ & (1,0,0), (0,1,0), (1,0,1), (0,1,1) \\
$x_3=1$ & (0,0,1), (0,1,1), (1,0,1), (1,1,1) & $z_3=1$ & (0,0,0), (1,1,0), (0,0,1), (1,1,1) \\
\hline
\end{tabular}
\end{center}
\caption{Resultant partitions from the transformation of Figure
\ref{coordFig}(b).} \label{coordPPTable}
\end{table}

Such transformations can be used to escape from local minima of
the Lagrangian. To see this consider a coordinate transformation
$\zeta$ from $\mathcal{X}$ to the new space $\mathcal{Z}$ such
that $\mathbf{z} = \zeta(\mathbf{x})$, and choose $q(\mathbf{z}) =
q(\mathbf{x})$ (i.e. do not change the associated probabilities).
Then the entropy contribution to the Lagrangian remains unchanged,
but the expected $G$ alters from $\sum_{\mathbf{x}} G(\mathbf{x})
q(\mathbf{x})$ to\endnote{The Jacobian factor is irrelevant as
$\zeta$ is a permutation.}
\begin{equation*}
\sum_{{\mathbf{x}}} G_{\mathcal{X}}({\mathbf{x}})
{q}(\mathbf{x}) \triangleq \sum_{{\mathbf{x}}} G(\zeta({\mathbf{x}}))
{q}(\mathbf{x}) = \sum_{\mathbf{z}}
G(\zeta\bigl(\mathbf{z})\bigr) q(\mathbf{z}).
\end{equation*}
This means that the gradient of the maxent Lagrangian will
typically differ before and after the application of $\zeta$. In
particular, what was a local minimum with zero gradient before the
semicoordinate transformation may not be a local minimum after the
transformation and the resultant shuffling of utility values. As
difficult problems typically have many local minima in their
Lagrangian, such semicoordinate transformations may prove very
useful.

\begin{figure}
\begin{center}
\mbox{
\subfigure[]{\includegraphics[width=0.75\textwidth]{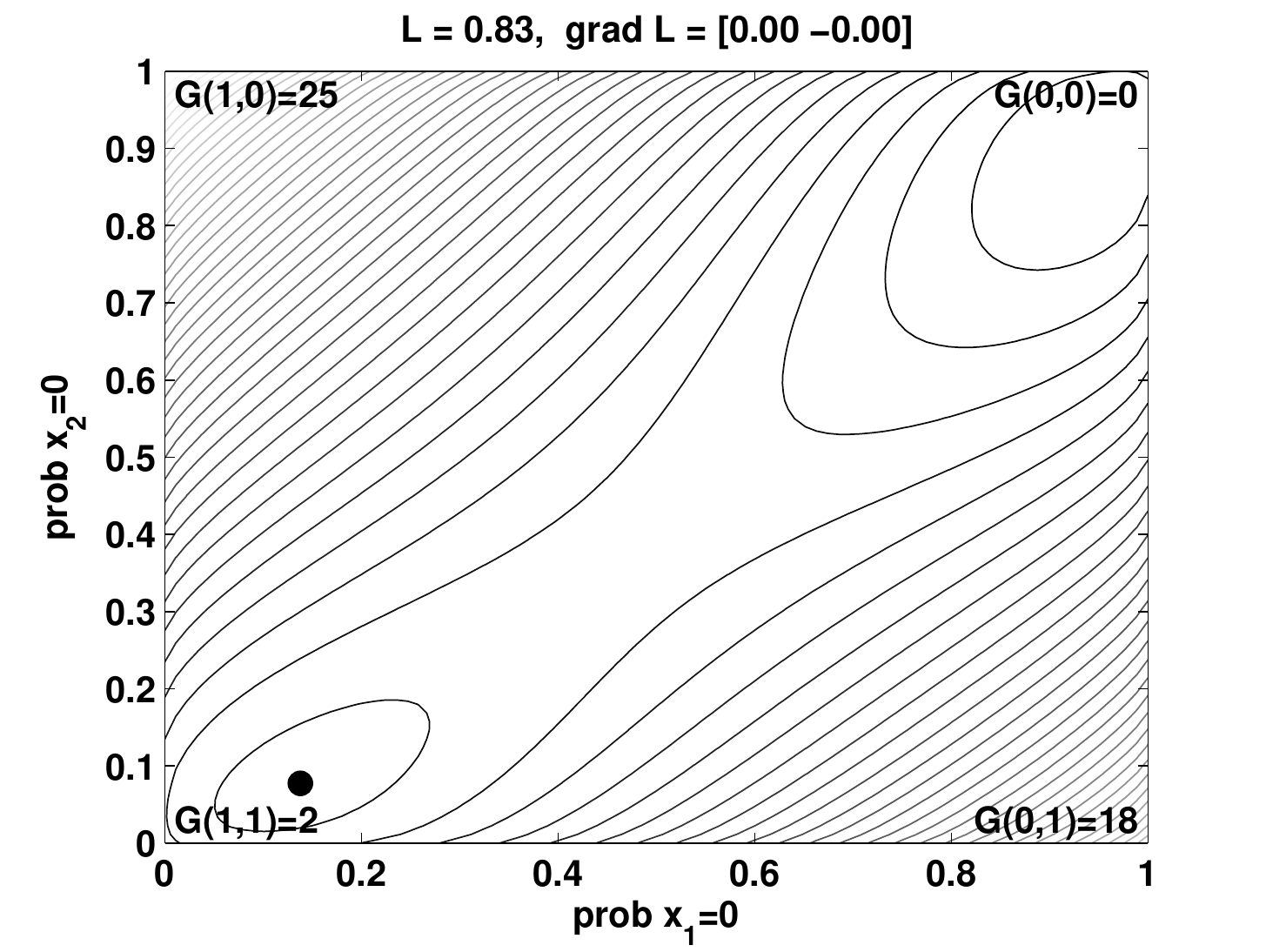}}}
\mbox{
\subfigure[]{\includegraphics[width=0.47\textwidth]{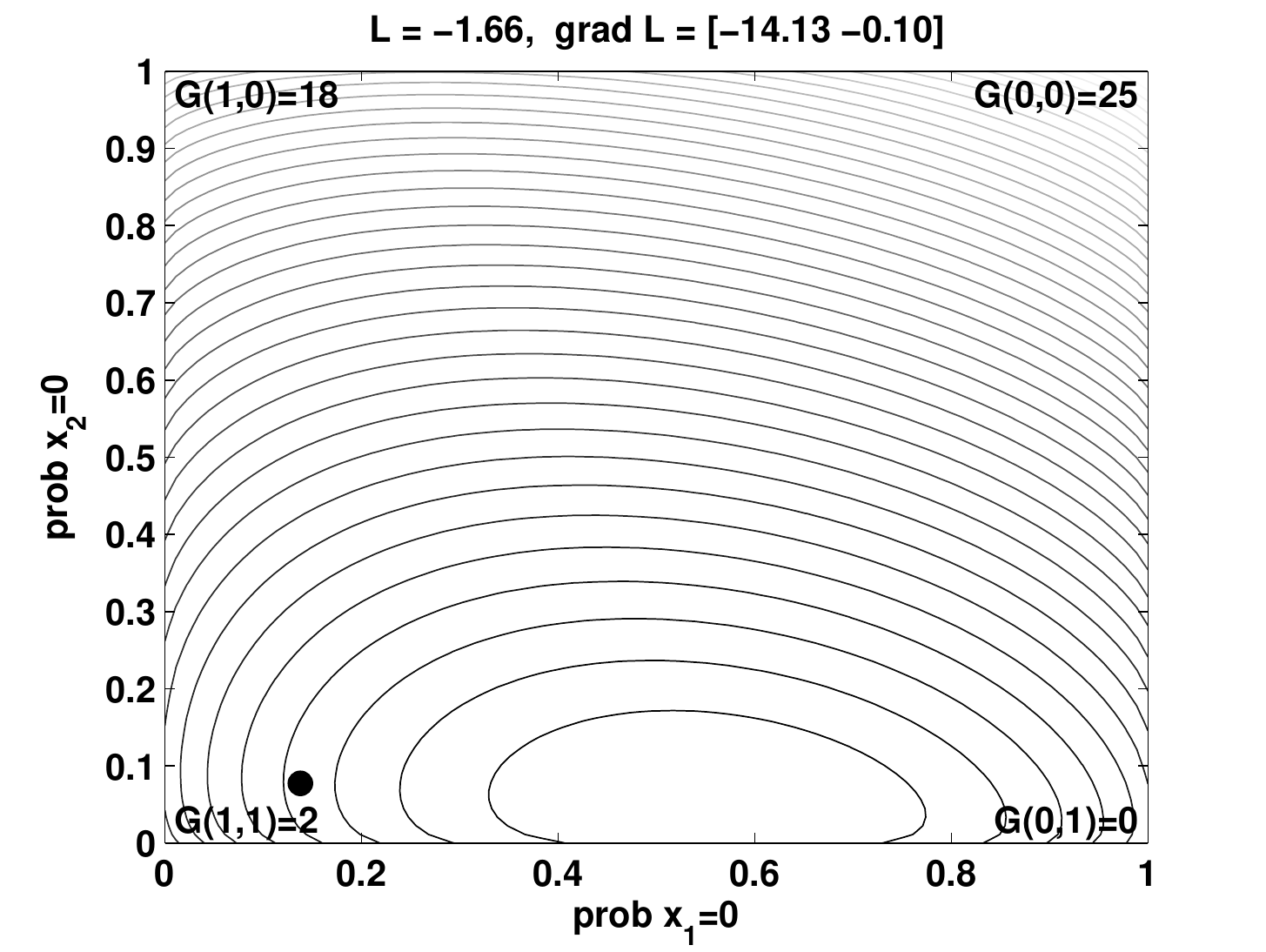}}}
\hspace{1mm} \mbox{
\subfigure[]{\includegraphics[width=0.47\textwidth]{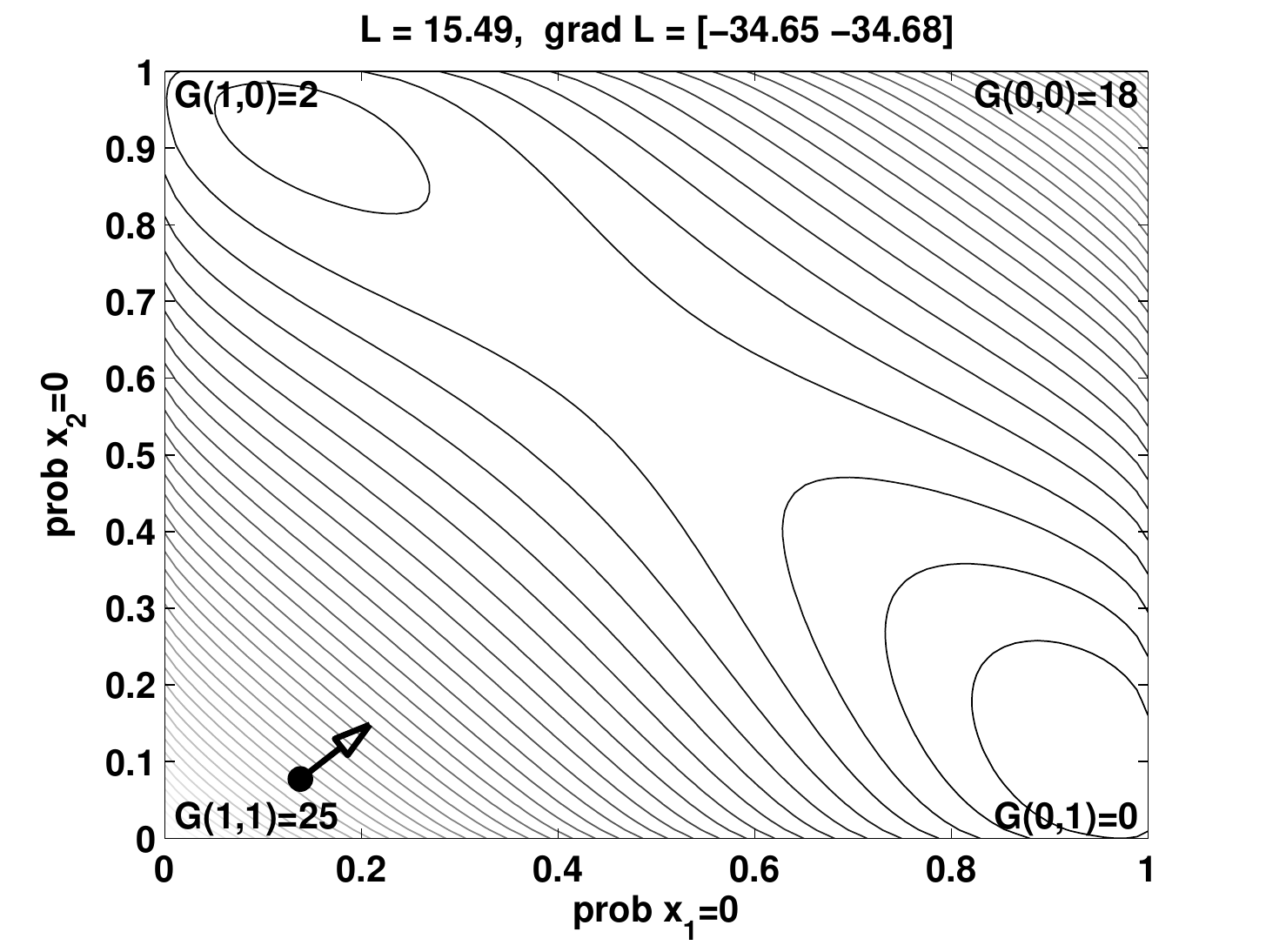}}}
\end{center}
\caption{(a) Original Lagrangian function. The suboptimal local
minimum is located at $q^*$ which is indicated with a dot in the
lower left corner. (b) The Lagrangian under the coordinate
transformation which minimizes $L(q^*)$. (c) The Lagrangian under
the coordinate transformation which maximizes the norm of the
gradient at $q^*$. The direction of the negative gradient is
indicated by a black arrow. } \label{localMinFig}
\end{figure}

A simple example is shown in \ref{localMinFig}(a) where a Lagrangian
surface for 2 binary variables is shown. The utility values are
$G(0,0)=0, G(1,0) = 25, G(0,1) = 18, G(1,1) = 2$. If the temperature
is 7 in units of the objective then the global minimum is at
$q_1(0)=0.95, q_2(0)=0.91$ where $L = -0.82$. At this temperature
there is a suboptimal local minimum (indicated by the dot in the
lower left) located at $q^*_1(0)=0.14, q^*_2(0)=0.08$ where
$L=0.83$.

There are a number of criteria that might be used to determine a
semicoordinate transformation to escape from this local minimum
$q^*$. Two simple choices are to select the transformation that
minimizes the new value of the maxent Lagrangian (i.e., minimize
$L(q^*)$), or to select the transformation which results in the
largest gradient of the maxent Lagrangian at $q^*$, (i,e,,
maximize $\| \nabla_q L(q^*)\|$). For this simple problem the
results of both these choices are shown as Figures
\ref{localMinFig}(b) and \ref{localMinFig}(c) respectively. The
transformation in each of these cases is determined by optimizing
over semicoordinate transformations (permutations), while keeping
the probabilities fixed, to either minimize the Lagrangian value
at $q^*$, or maximize the norm of the gradient at $q^*$.

In principle, this semicoordinate search can be embedded within
any optimization to dynamically escape local minima as they are
encountered. Importantly, the search criteria listed above require
no look ahead in order to identify the best semicoordinate
permutation. However, in practice, since we can not search over
arbitrarily large permutation spaces we must select a few of the
variables and permute amongst their possible joint
moves.\endnote{Alternatively, we can parameterize a smaller space
of candidate permutations and select the best from amongst this
candidate set.} By composing such permutations we can easily
account for the escape from multiple local minima. Heuristics for
the selection of ``permuting" variables, and the results of this
procedure await future work.

\subsection{Semicoordinate Transformations for Mixture Distributions}

In this section we turn to a different use of semicoordinate
transformations. We have previously described how the Lagrangian
measuring the distance of a product distribution to a Boltzmann
distribution may be minimized in a distributed fashion. We now
extend these results to mixtures of product distributions in order
to represent multiple product distribution solutions at once. We
can always do that by means of a semicoordinate transformation of
the underlying variables. In this section we demonstrate this
explicitly.

Let $\mathcal{X}$ indicate the set of variables in a space of
dimension $d_{\mathcal{X}}$, and let $\mathcal{Z}$ be the original
(pre-transformation) $n$-dimensional space over which $G$ is
defined. We identify a product distribution over $\mathcal{X}$
(with $d_{\mathcal{X}}> n$), and an appropriately chosen mapping
$\pmb{\zeta} : \mathcal{X} \rightarrow \mathcal{Z}$ which induce a
mixture distribution over $\mathcal{Z}$.

To see this consider an $M$ component mixture distribution over
$n$ variables, which we write as: $q(z) = \sum_{m=1}^M q^0(m)
q^m(\mathbf{z})$ with $\sum_{m=1}^M q^0(m)=1$ and $q^m(\mathbf{z})
= \prod_{i=1}^n q^m_i(z_i)$. We express this $q(\mathbf{z})$ as
(the image of) a product distribution over a space $\mathcal{X}$
of dimension $d_{\mathcal{X}} = 1+Mn$. Intuitively, the first
dimension of $\mathcal{X}$ (indicated as $x^0\in[1,M]$) labels the
mixture, and the remaining $Mn$ dimensions (indicated as $x^m_i\in
\mathcal{Z}_i$) correspond to each of the original $n$ dimensions
for each of the $M$ mixtures.

More precisely, write out the $\mathcal{X}$-space product
distribution as $q_\mathcal{X}(\mathbf{x}) = q^0(x^0)\prod_{m=1}^M
q^m(\mathbf{x}^m)$ with $q^m(\mathbf{x}^m) = \prod_{i=1}^n
q^m_i(x^m_i)$ for $\mathbf{x} =
[x^0,\mathbf{x}^1,\cdots,\mathbf{x}^M]$ and $\mathbf{x}^m =
[x^m_1, \cdots, x^m_n]$. The density in $\mathcal{X}$ and
$\mathcal{Z}$ are related as usual by $q(\mathbf{z}) =
\sum_\mathbf{x} q_{\mathcal{X}}(\mathbf{x}) \delta\bigl(\mathbf{z}
- \zeta(\mathbf{x})\bigr)$ with the delta function acting on
vectors being understood component-wise. If we label the
components of $\zeta$ so that
$z_i=\zeta_i(x^0,\mathbf{x}^1,\cdots,\mathbf{x}^M) \triangleq
x_i^{x^0}$ we find
\begin{align*}
q(\mathbf{z}) &= \sum_{x^0} q^0(x^0) \sum_{\mathbf{x}^1, \cdots,
\mathbf{x}^M} \prod_m q^m(\mathbf{x}^m) \prod_i
\delta\bigl(z_i-\zeta_i(x^0,\mathbf{x}^1,\cdots,\mathbf{x}^M)\bigr)
\\ &= \sum_{x^0} q^0(x^0) \sum_{\mathbf{x}^1,\cdots,\mathbf{x}^M}
\prod_m q^m(\mathbf{x}^m) \prod_i \delta\bigl(z_i-x^{x^0}_i\bigr)
\\ &= \sum_{x^0} q^0(x^0) \sum_{\mathbf{x}^{x^0}}
q^{x^0}(\mathbf{x}^{x^0}) \prod_i \delta\bigl(z_i-x^{x^0}_i\bigr)
\\ &= \sum_{x^0} q^0(x^0) q^{x^0}(\mathbf{z})
\end{align*}
Thus, under $\zeta$ the product distribution $q_{\mathcal{X}}$ is
mapped to the mixture of products $q(\mathbf{z}) = \sum_m
q^0(m)q^m(\mathbf{z})$ (after relabelling $x^0$ to $m$), as desired.

The maxent Lagrangian of the $\mathcal{X}$ product distribution
$q_\mathcal{X}(\mathbf{x})$ is
$$
\mathcal{L}_{\mathcal{X}}(q_\mathcal{X}) = \sum_m q^0(m)
\mathbb{E}_{q^m}(G) - T \Bigl[ S(q^0) + \sum_{m=1}^M S(q^m)\Bigr].
$$
This Lagrangian contains a term pushing us (as we search for the
minimizer of that Lagrangian) to maximize the entropy of the
mixture weights, but it provides no incentive for the
distributions $q^m$ to differ from each other. As we have argued,
it is desirable to have the different mixtures capture different
solutions, and so we modify the Lagrangian function slightly.

If we wish the $q^m$ differ from one another, we instead consider
the maxent Lagrangian defined over $q(\mathbf{z})$. In this case
\begin{align*}
\mathcal{L}_{\mathcal{Z}}(q) &= \sum_\mathbf{z} G\bigl(\mathbf{z}) q(\mathbf{z})
- T S(q) = \sum_\mathbf{x} G\bigl(\pmb{\zeta}(\mathbf{x})\bigr)
q_\mathcal{X}(\mathbf{x}) - T S(q) \\ &= \sum_m q^0(m)
\mathbb{E}_{q^m}(G) - T S\biggl(\sum_{m} q^0(m)
q^{m}(\mathbf{z})\biggr).
\end{align*}
The entropy term differs crucially in these two maxent
Lagrangians. To see this add and subtract $T\sum_m q^0(m) S(q^m)$
to the $\mathcal{Z}$ Lagrangian to find
\begin{equation}
\mathcal{L}_{\mathcal{Z}}(q) = \sum_m q^0(m) \mathcal{L}(q^m) - T J(q)
\label{jointLEq}
\end{equation}
where each $\mathcal{L}(q^m)$ is a maxent Lagrangian as given by Eq.
\eqref{freeEnergyEq}, and $J(q)\ge 0$ is a modified version of the
Jensen-Shannon (JS) distance,
\begin{equation*}
J(q) = S\Bigl(\sum_m q^0(m) q^m\Bigr) -\sum_m q^0(m) S(q^m) =
-\sum_m \sum_{\mathbf{z}} q^0(m) q^m(\mathbf{z}) \ln
\frac{q(\mathbf{z})}{q^m(\mathbf{z})}.
\end{equation*}
Conventional Jensen-Shannon distance is defined to compare two
distributions to each other, and gives those distributions equal
weight. In contrast, the generalized JS distance $J(q)$ concerns
multiple distributions, and weights them nonuniformly, according
to $q^0(m)$.

$J(q)$ is maximized when the $q^m$ are all different from each
other. Thus, its inclusion into the Lagrangian pushes the mixing
components away (in $\mathcal{X}$) from one another.  In this, we
can view Eq.~\eqref{jointLEq} as a novel derivation of (a
generalized version of) Jensen-Shannon distance.  Unfortunately,
the JS distance also couples all of the variables (because of the
sum inside the logarithm) which prevents a highly distributed
solution.

To address this, in this paper we replace $J(q)$ in
$\mathcal{L}_{\mathcal{Z}}(q)$ with another function which
lower-bounds $J(q)$ but which requires less communication between
agents. It is this modified Lagrangian that we will minimize.

\subsection{A Variational Lagrangian}

Following \cite{jj98}, we introduce $M$ variational functions
$w(\mathbf{z}|m)$ and lower-bound the true JS distance. We begin
with the identity
\begin{align*}
J(q) &= -\sum_m \sum_{\mathbf{z}} q^0(m) q^m(\mathbf{z}) \ln
\left[\frac{1}{w(\mathbf{z}|m)} q^0(m) \frac{w(\mathbf{z}|m)
q(\mathbf{z})}{q^0(m) q^m(\mathbf{z})} \right]
\\ &= \sum_m \sum_{\mathbf{z}} q^0(m) q^m(\mathbf{z}) \ln
w(\mathbf{z}|m)) - \sum_m q^0(m)\ln q^0(m) \\ &\phantom{=} -
\sum_m \sum_{\mathbf{z}} q^0(m) q^m(\mathbf{z}) \ln
\frac{w(\mathbf{z}|m) q(\mathbf{z})}{q^0(m) q^m(\mathbf{z})}.
\end{align*}
Now replace $M$ of the $-\ln$ terms with the lower
bound $-\ln z \ge -\nu z + \ln \nu + 1$ obtained from the Legendre
dual of the logarithm to find
\begin{align*}
J(q) \ge J(q,w,\nu) &\triangleq \sum_m \sum_{\mathbf{z}} q^0(m)
q^m(z) \ln w(\mathbf{z}|m) - \sum_m q^0(m)\ln q^0(m) \\
& \phantom{\ge} - \sum_m \nu_m \sum_{\mathbf{z}}
w(\mathbf{z}|m)q(z) + \sum_m q^0(m) \ln \nu_m + 1.
\end{align*}
Optimization over $w$ and $\nu$ maximizes this lower bound. To
further aid in distributing the algorithm we restrict the class of
variational $w(\mathbf{z}|m)$ to products: $w(\mathbf{z}|m) =
\prod_i w_i(z_i|m)$. For this choice
\begin{align}
J(q,w,\nu)  &\triangleq  \sum_m q^0(m) \left\{B^{m,m} -
\sum_{\tilde{m}} A^{m,\tilde{m}} \nu_{\tilde{m}} + \ln \nu_m
\right\} + S(q^0) + 1 \label{JEq}
\end{align} where
$A^{\tilde{m},m}_i \triangleq \sum_{z_i} q^{\tilde{m}}_i(z_i)
w_i(z_i|m)$, $A^{\tilde{m},m} \triangleq \prod_{i=1}^d
A^{\tilde{m},m}_i$, $B^{m,m}_i \triangleq \sum_{z_i} q^m_i(z_i)\ln
w_i(z_i|m)$, and $B^{m,m} \triangleq \sum_{i=1}^d
B^{m,m}_i$.\endnote{Note that if $w_i(z_i|m) = 1/|\mathcal{Z}_i|$
is uniform across $z_i$ then $A^{\tilde{m},m}_i =
1/|\mathcal{Z}_i|$ and $B^{m,m}_i = -\ln |\mathcal{Z}_i|$.
Maximizing over $\nu_m$ we find that
$J(q,w=1/|\mathcal{Z}|,\nu=\nu^*) = 0$. Thus, maximizing with
respect to $w$ increases the JS distance from 0.} At any
temperature $T$ the variational Lagrangian which must be minimized
with respect to $q$, $w$ and $\nu$ (subject to appropriate
positivity and normalization constraints) is then
\begin{equation}
\mathcal{L}_{\mathcal{Z}}(q,w,\nu) = \sum_m q^0(m)
\mathcal{L}(q^m) - T J(q,w,\nu) \label{mixLagrangeEq}
\end{equation}
with $J(q,w,\nu)$ given by Eq. \eqref{JEq}.

\section{Minimizing the Mixture Distribution Lagrangian}\label{minimizeSect}

Equating the gradients with respect to $w$ and $\nu$ to zero gives
\begin{gather}
\frac{1}{\nu_m} = \frac{1}{q^0(m)} \sum_{\tilde{m}} q^0(\tilde{m})
A^{\tilde{m},m}. \label{nuEq} \\ w_i(z_i|m) \propto \frac{q^0(m)
q^m_i(z_i)}{\nu_m} \left[\sum_{\tilde{m}} q^0(\tilde{m})
q^{\tilde{m}}_i(z_i)
\frac{A^{\tilde{m},m}}{A^{\tilde{m},m}_i}\right]^{-1}. \label{wEq}
\end{gather}
The dependence of $\mathcal{L}_{\mathcal{Z}}$ on $q^0(m)$ is
particularly simple: $\mathcal{L}_{\mathcal{Z}}(q,w,\nu) \approx
\sum_m q^0(m) \mathcal{E}(m) - T \bigl(S(q^0) +1\bigr)$ up to
$q^0$-independent terms where
\begin{equation*}
\mathcal{E}(m) \triangleq \mathbb{E}_{q^m}(G) -T \left( S[q^m] +
B^{m,m} - \sum_{\tilde{m}} A^{m,\tilde{m}} \nu_{\tilde{m}} + \ln
\nu_m \right),
\end{equation*}
Thus, the mixture weights are Boltzmann
distributed with energy function $\mathcal{E}(m)$:
\begin{equation}
q^0(m) = \frac{\exp\bigl(-\mathcal{E}(m)/T\bigr)}{\sum_{\tilde{m}}
\exp\bigl(-\mathcal{E}(\tilde{m})/T\bigr)}. \label{q0Eq}
\end{equation}
The determination of $q^m_i(z_i)$ is similar. The relevant terms
in $\mathcal{L}_{\mathcal{Z}}$ involving $q^m_i(z_i)$ are
$\mathcal{L}_{\mathcal{Z}} \approx q^0(m) \sum_{z_i}
\mathcal{E}_m(z_i) q^m_i(z_i) - T S(q^m_i)$ where
\begin{equation*}
\mathcal{E}_m(z_i) = \mathbb{E}_{q^m_{- i}}(G|z_i) -
T\biggl(\ln w_i(z_i|m) - \sum_{\tilde{m}}
\frac{A^{m,\tilde{m}}}{A^{m,\tilde{m}}_i} \nu_{\tilde{m}}
w_i(z_i|\tilde{m}) \biggr).
\end{equation*}
As before, the conditional expectation $\mathbb{E}_{q^m_{-
i}}(G|z_i)$ is $\sum_{z_{- i}} G(z_i,\mathbf{z}_{- i})q^m_{-
i}(\mathbf{z}_{- i})$. The mixture probabilities are thus
determined as
\begin{equation}
q^m_i(z_i) =
\frac{\exp\bigl(-\mathcal{E}_m(z_i)/T\bigr)}{\sum_{z_i}
\exp\bigl(-\mathcal{E}_m(z_i)/T\bigr)}. \label{qmEq}
\end{equation}

\subsection{Agent Communication}
As desired these results require minimal communication between
agents. A supervisory agent, call this the 0-agent, is assigned to
manage the determination of $q^0(m)$, and $(i,m)$-agents manage
the determination of $q^m_i(z_i)$. The $M$ $(i,m)$-agents for a
fixed $i$ communicate their $w_i(z_i|m)$ to determine
$A^{m,\tilde{m}}_i$. These results along with the $B^{m,m}_i$ from
each $(i,m)$ agent are then forwarded to the 0-agent who forms
$A^{m,\tilde{m}}$ and $B^{m,m}$ broadcasts this back to all
$(i,m)$-agents. With these quantities and the local estimates for
$\mathbb{E}_{q^m_{-i}}(G|z_i)$, all $q_i^m$ can be updated
independently.

\section{Experiments}

In this section we demonstrate our methods on some simple
problems. These examples are illustrative only, and for further
examples the reader is directed to
\cite{biwo04a,biwo04b,biwo04c,lewo04b,wole04} for related
experiments.

We test the the mixture semicoordinate probability collective
method on two different problems: a $k$-sat constraint
satisfaction problem having multiple feasible solutions, and
optimization of an unconstrained optimization of an $NK$ function.

\subsection{k-sat} \label{ksat}

The $k$-sat problem is perhaps the best studied CSP \cite{mepa02}.
The goal is to assign $N$ binary variables (labelled $z_i$)
true/valse values so that $C$ disjunctive clauses are satisfied.
The $a$th clause involves $k$ variables labelled by $v_{a,j}\in
[1,N]$ (for $j\in[1,k]$), and $k$ binary values associated with
each $a$ and labelled by $\sigma_{a,j}$. The $a$th clause is
satisfied iff $\bigvee_{j=1}^k [z_{v_{a,j}}=\sigma_{a,j}]$ so we
define the $a$th constraint as
\begin{equation*}
c_a(\mathbf{z}) = \begin{cases} 0 & \text{if} \; \bigvee_{j=1}^k
[z_{v_{a,j}}=\sigma_{a,j}] \\ 1 & \text{otherwise}
\end{cases}.
\end{equation*}
As the $a$th clause is violated only when all $z_{v_{a,j}} =
\overline{\sigma}_{a,j}$ (with $\overline{\sigma}$ defined to be
$\text{not} \, \sigma$), the Lagrangian over product distributions
can be written as $\mathcal{L}(q) = \pmb{\lambda}^\top
\mathbf{c}(q) - TS(q)$ where $\mathbf{c}(q)$ is the $C$-vector of
expected constraint violations, and $\pmb{\lambda}$ is the $C$
vector of Lagrange multipliers. The $a$'th component of
$\mathbf{c}(q)$ is the expected violation of the $a$th clause, and
is given by $c_a({q}) \triangleq \sum_{\mathbf{z}} c_a(\mathbf{z})
q(\mathbf{z}) = \prod_{j=1}^k
q_{v_{a,j}}(\overline{\sigma}_{a,j})$. Note that no Monte Carlo
estimates are required to evaluate this quantity. Further, the
only communication required to evaluate $G$, and its conditional
expectations is between agents appearing in the same clause.
Typically, this communication network is sparse; for the $N=100$,
$k=3$, $C=430$ variable problem we consider here each agent
interacts with only $6$ other agents on average.

\begin{figure}
\begin{center} \mbox{
\subfigure[]{\includegraphics[width=0.45\textwidth]{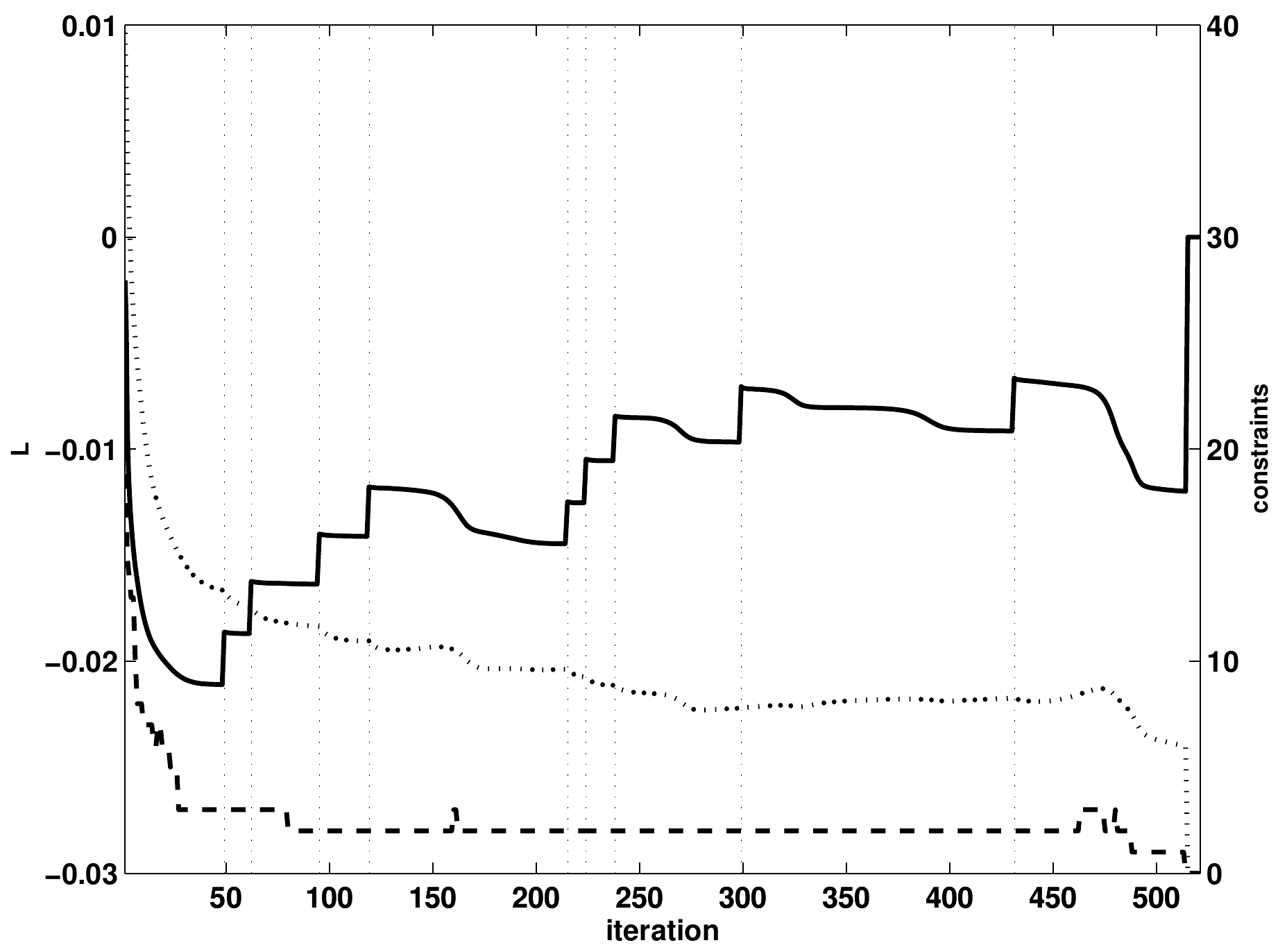}}}
\hspace{3mm}
\mbox{\subfigure[]{\includegraphics[width=0.4\textwidth]{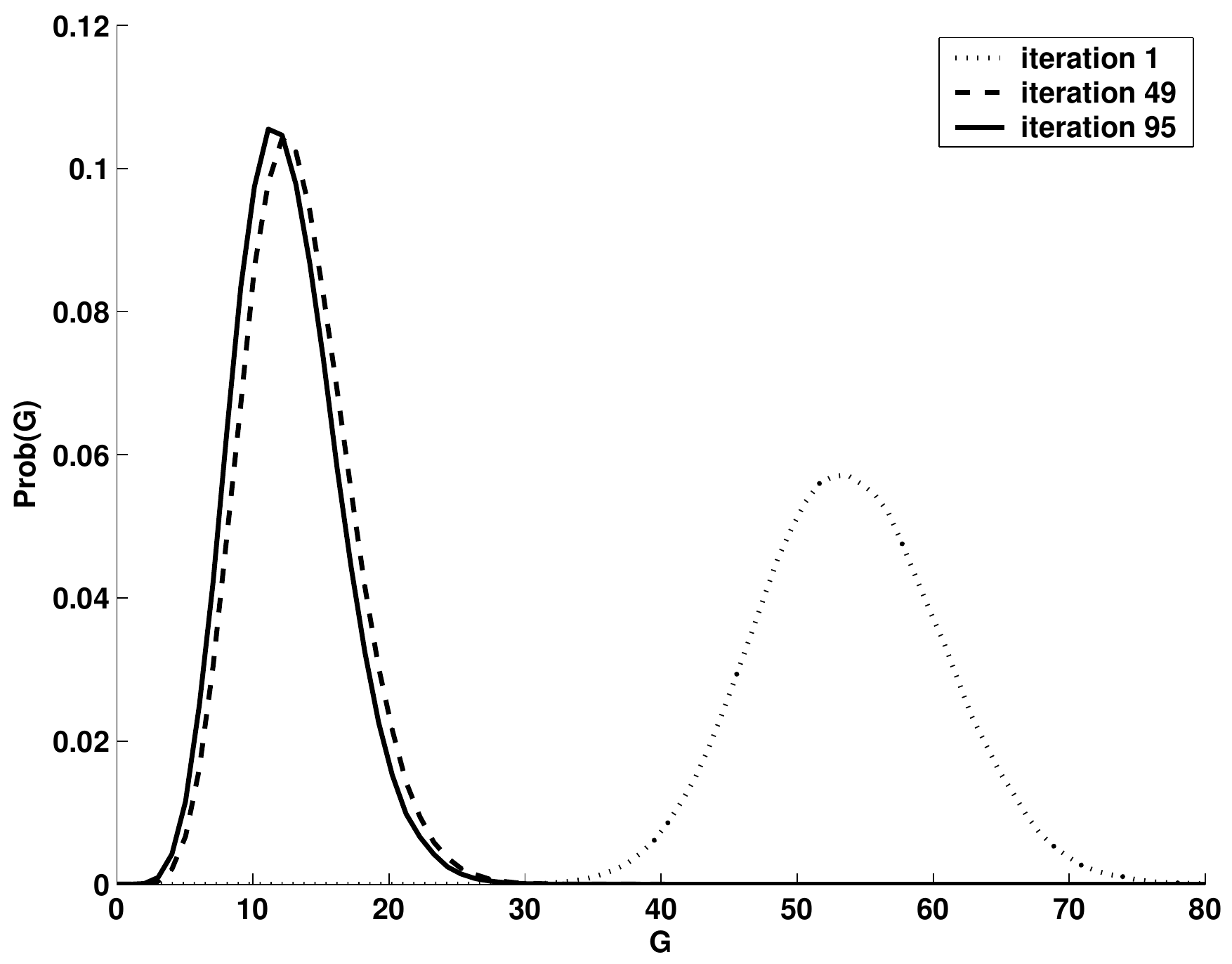}}}
\end{center}
\caption{(a) Evolution of Lagrangian value (solid line), expected
constraint violation (dotted line), and constraint violations of
most likely configuration (dashed line). (b) ${P}(G)$ after
minimizing the Lagrangian for the first 3 multiplier settings. At
termination $P(G) = \delta(G)$.} \label{100Fig}
\end{figure}

\begin{figure}
\centering{\includegraphics[width=0.5\textwidth]{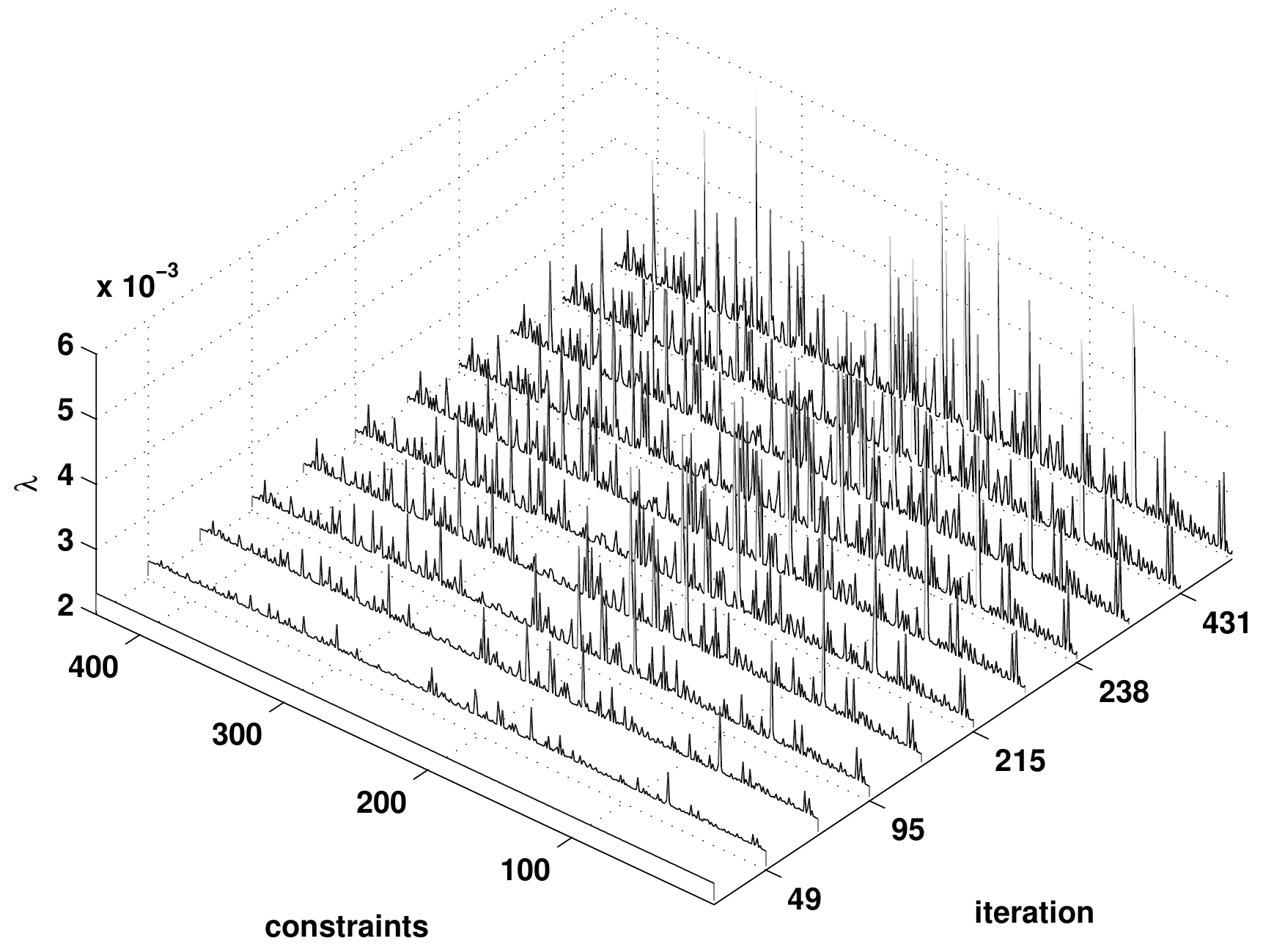}}
\caption{Each constraint's Lagrange multiplier versus the
iterations when they change.} \label{timePlotFigb}
\end{figure}

We first present results for a single product distribution. For
any fixed setting of the Lagrange multipliers, the Lagrangian is
minimized by iterating Eq. \eqref{nearNewtEq}. If the minimization
is done by the Brouwer method, any random subset of variables, no
two of which appear in the same clause, can be updated
simultaneously. This eliminates ``thrashing" and ensures that the
Lagrangian decreases at each iteration.

The minimization is terminated at a local minimum ${q}^*$ which is
detected when the norm of the gradient falls below a threshold. If
all constraints are satisfied at ${q}^*$ we return the solution
$\mathbf{z}^* = \argmax_{\mathbf{z}} q^*(\mathbf{z})$, otherwise
the Lagrange multipliers are updated according to Eq.
\eqref{eq:updateconstrain}. In the present context, this updating
rule offers a number of benefits. Firstly, those constraints which
are violated most strongly have their penalty increased the most,
and consequently, the agents involved in those constraints are
most likely to alter their state. Secondly, the Lagrange
multipliers contain a history of the constraint violations (since
we keep adding to $\pmb{\lambda}$) so that when the agents
coordinate on their next move they are unlikely to return a
previously violated state. This mimics the approach used in taboo
search where revisiting of configurations is explicitly prevented,
and aids in an efficient exploration of the search space. Lastly,
rescaling the Lagrangian after each update of the multipliers by
$\pmb{1}^\top \pmb{\lambda} \triangleq \sum_a \lambda_a$ gives
$\mathcal{L}({q}) = \hat{\pmb{\lambda}}{}^\top \mathbf{c}({q}) -
\hat{T} S({q})$ where $\hat{\pmb{\lambda}} =
\pmb{\lambda}/\mathbf{1}^\top \pmb{\lambda}$ and $\hat{T} = T /
\mathbf{1}^\top \pmb{\lambda}$. Since $\sum_a \hat{\lambda}_a=1$
the first term reweights clauses according to their expected
violation, while the temperature $\hat{T}$ cools in an automated
annealing schedule as the Lagrange multipliers increase. Cooling
is most rapid when the expected constraint violation is large and
slows as the optimum is approached. The parameters
$\alpha_\lambda^t$ thus govern the overall rate of cooling. We
used the fixed value $\alpha_\lambda^t =0.5$ in the reported
experiments.

Figure \ref{100Fig} presents results for a 100 variable $k=3$
problem using a single mixture. The problem is satisfiable formula
\texttt{uf100-01.cnf} from SATLIB (\url{www.satlib.org}). It was
generated with the ratio of clauses to variables being near the
phase transition, and consequently has few solutions. Fig.
\ref{100Fig}(a) shows the variation of the Lagrangian, the
expected number of constraint violations, and the number of
constraints violated in the most probable state ${z}_{\text{mp}}
\triangleq \argmax_{\mathbf{z}} q(\mathbf{z})$ as a function of
the number of iterations. The starting state is the maximum
entropy configuration of uniform $q_i$, and the starting
temperature is $T=1.5\cdot 10^{-3}$. The iterations at which the
Lagrange multipliers are updated are indicated by vertical dashed
lines which are clearly visible as discontinuities in the
Lagrangian values. To show the stochastic underpinnings of the
algorithm we plot in Fig. \ref{100Fig}(b) the probability density
of the number of constraint violations obtained as $P(G) =
\sum_{\mathbf{z}} q(\mathbf{z}) \delta\bigl(G - \sum_a
c_a(\mathbf{z}) \bigr)$.\endnote{In determining the density $10^4$
samples were drawn from $q(\mathbf{z})$ with Gaussians centered at
each value of $G(\mathbf{z},{\bf{1}})$ and with the width of all
Gaussians determined by cross validation of the log likelihood.
The fact that there is non-zero probability of obtaining
non-integral numbers of constraint violations is an artifact of
the finite width of the Gaussians.} We see the downward movement
of the expected value of $G$ as the algorithm progresses. Figure
\ref{timePlotFigb} shows the evolution of the renormalized
Langrange multipliers $\hat{\pmb{\lambda}}$. At the first
iteration the multiplier for all clauses are equal. As the
algorithm progresses weight is shifted amongst difficult to
satisfy clauses.

\begin{figure}
\begin{center}
\mbox{
\subfigure[]{\includegraphics[width=0.475\textwidth]{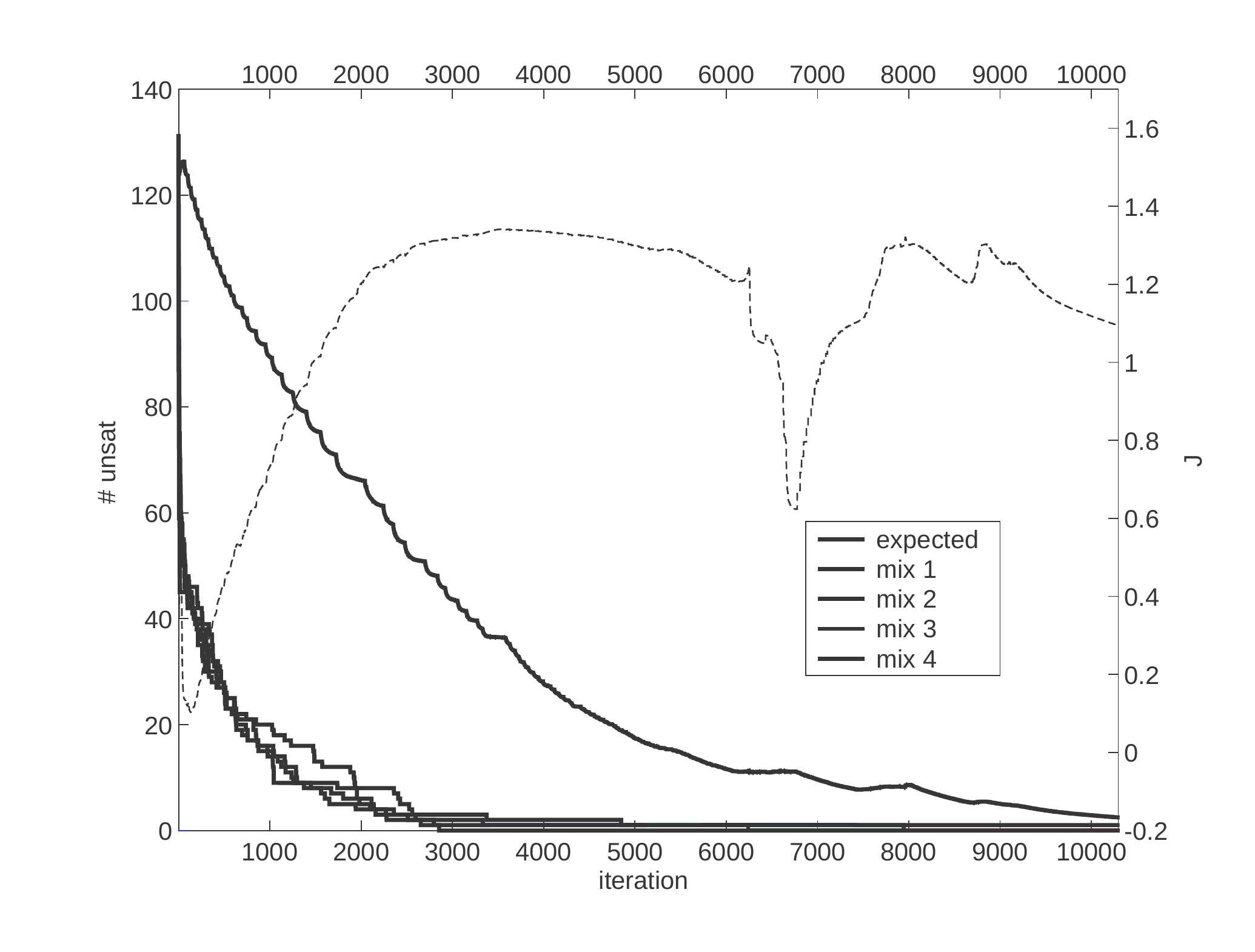}}}
\hspace{2mm}
\mbox{\subfigure[]{\includegraphics[width=0.475\textwidth]{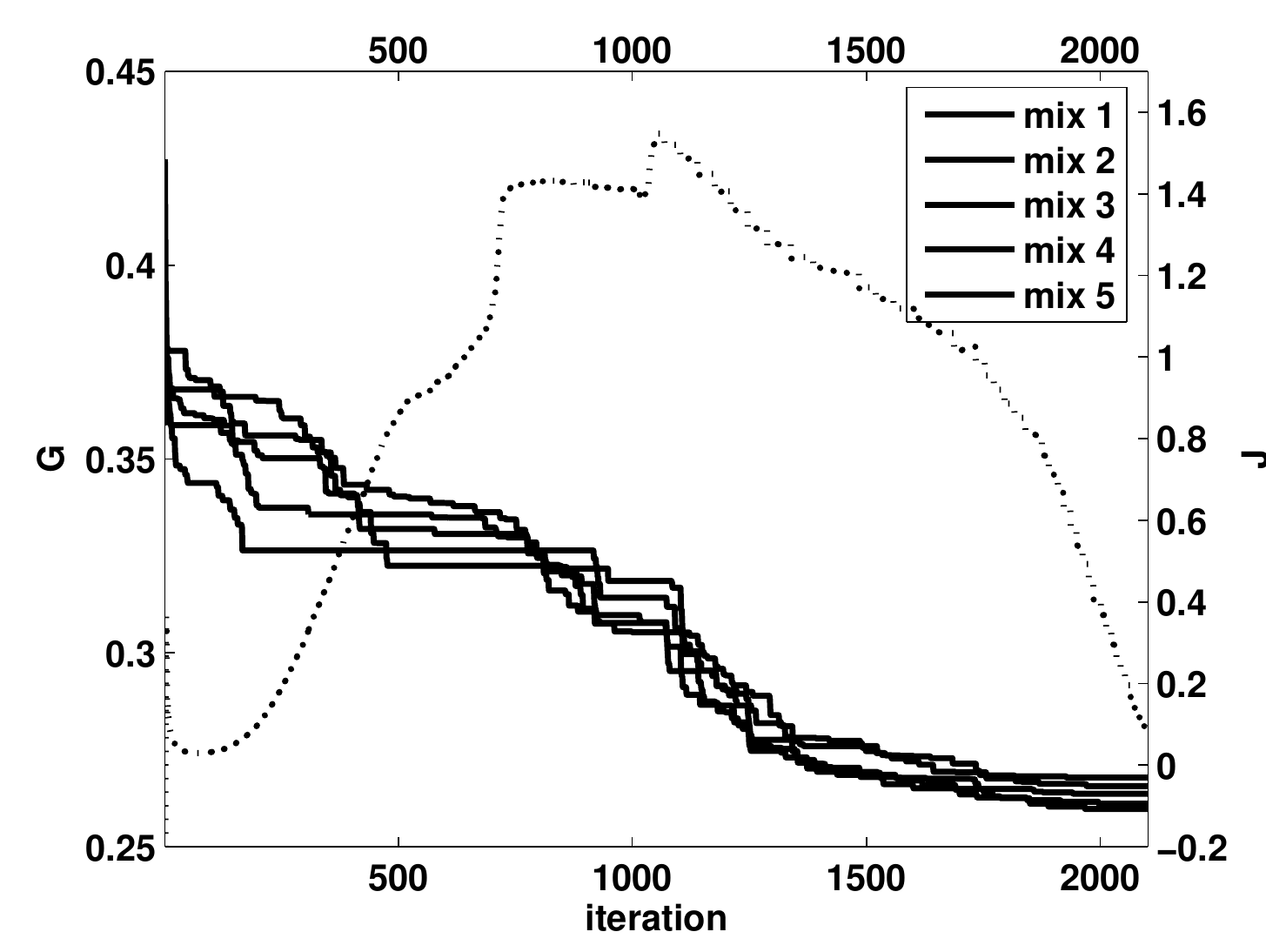}}}
\end{center}
\caption{(a) The solid curves show the number of unsatisfied clauses
in the most probable configuration $z_{\text{mp}}$ of each of the 4
mixtures vs iterations. The topmost solid black line plots the
expected number of violations, and the dashed black line shows the
approximation to the JS distance. (b) The solid curves show the
evolution of the $G$ value of the best $z_{\text{mp}}$
configurations for each of 5 mixtures versus number of iterations.
The dashed black line shows the corresponding approximation to the
JS distance.} \label{mixFig}
\end{figure}

Results on a larger problem with multiple mixtures are shown in
Fig. \ref{mixFig}(a). This is the 250 variable/1065 clause problem
\texttt{uf250-01.cnf} from SATLIB with the first 50 clauses
removed so that the problem has multiple solutions. The
optimization is performed by selecting a random subset of
variables, no two of which appear in the same clause, at each
iteration, and updating according to Eqs. \eqref{nuEq},
\eqref{wEq}, \eqref{q0Eq}, and \eqref{qmEq}. After convergence to
a local minimum the Lagrange multipliers are updated as above
according to the expected constraint violation. The initial
temperature is $10^{-1}$. We plot the number of constraints
violated in the most probable state of each mixture as a function
of the number of updates. as well as the expected number of
violated constraints. After 8000 steps three distinct solutions
are found along with a fourth configuration which violates a
single constraint.

\subsection{Minimization of NK Functions} \label{nk}

Next we consider an unconstrained discrete optimization problem.
The $NK$ model defines a family of tunably difficult optimization
problems \cite{KLNK}. The objective over $N$ binary variables
$\{z_i\}_{i=1}^N$ is defined as the average of $N$ randomly
generated contributions depending on $z_i$ and $K$ other randomly
chosen variables $z_i^1\cdots z_i^K$: $G(\mathbf{z}) = N^{-1}
\sum_{i=1}^N E_i(z_i; z_i^1, \cdots z_i^K)$. For each of the
$2^{K+1}$ local configurations $E_i$ is assigned a value drawn
uniformly from 0 to 1. $K$ which ranges between 0 and $N-1$
controls the number of local minima; under Hamming neighborhoods
$K=0$ optimization landscapes have a single global optimum and
$K=N-1$ landscapes have on average $2^N/(N+1)$ local minima.
Further properties of $NK$ landscapes may be found in
\cite{durrett.limic.ea:rigorous}. Fig. \ref{mixFig}(b) plots the
energy of a 5 mixture model for a multi-modal $N=300$ $K=2$
function. The $K-1$ spins other than $i$ upon which $E_i$ depends
were selected at random. At termination of the PC algorithm (at a
small but non-zero temperature), five distinct configurations are
obtained with the nearest pair of solutions having Hamming
distance 12. Note that unlike the $k$-sat problem which has
multiple configurations all having the same global minimal energy,
the JS distance (the dashed curve) of Fig. \ref{mixFig}(b) drops
to zero as the temperature decreases. This is because at exactly
zero temperature there is no term forcing different solutions, and
the Lagrangian is minimized by having all mixtures converge to
delta functions at the lowest objective value configuration. The
effect of mixtures enables the algorithm to simultaneously explore
multiple local extema before converging on the lowest objective
solution.

\section{Relation of PC to other work}

There has been much work from many fields related to PC. The maxent
Lagrangian has been used in statistical physics for over a century
under the rubric of ``free energy''. Its derivation in terms of
information theoretic statistical inference was by Jaynes
\cite{jayn57}. The maxent Lagrangian has also appeared occasionally
in game theory as a heuristic, without a statistical inference
justification (be it information-theoretic or otherwise)
\cite{fule98,shar04}.\endnote{However an attempt at a
first-principles derivation can be found in \cite{megi76}.}  In none of
this earlier work is there an appreciation for its relationship with
the related work in other fields.

In the context of distributed control/optimization, the
distinguishing feature of PC is that it does not view the variable
${\mathbf{x}}$ as the fundamental object, but rather the
distribution across it, $q$. Samples of that distribution are not
the direct object of interest, and in fact are only used if
necessary to estimate functionals of $q$. The fundamental
objective function is stated in terms of $q$.  As explicated in
the next subsection, the associated optimization algorithms are
related to some work in several fields. Heretofore those fields
have been unaware of each other, and of the breadth of their
relation to information theory and game theory.

Finally, we note that the maxent or $qp$ Lagrangian
$\mathcal{L}(q) = E_q(G) - TS(q)$ can be viewed as a
barrier-function (interior point) method with objective $E_q(G)$.
An entropic barrier function is used to enforce the constraints
$q_i(x_i) \ge 0 \; \forall i$ and $x_i$, with the constraint that
all $q_i$ sum to 1 being implicit.

\subsection{Various schemes for updating $q$}

We have seen that the $qp$ Lagrangian is minimized by the product
distribution $q$ given by
\begin{equation}
q_i(x_i) \propto \exp\bigl(-E_{q_{-i}}(G | x_i)/T\bigr) .
\label{eq:boltzG}
\end{equation}
Direct application of these equations that minimize the Lagrangian
form the basis of the Brouwer update rules. Alternatively,
steepest descent of the maxent Lagrangian forms the basis of the
Nearest Newton algorithm. These update rules have analogues in
conventional (non-PC) optimization. For example, Nearest-Newton is
based on Newton's method, and Brouwer updating is similar to
block-relaxation. This is one of the advantages of embedding the
original optimization problem involving $x$ in a problem involving
distributions across $x$: it allows us to solve problems over
non-Euclidean (e.g., countable) spaces using the powerful methods
already well-understood for optimization over Euclidean spaces.

However there are other PC update rules that have no direct
analogue in such well-understood methods for Euclidean space
optimization. Algorithms may be developed that minimize the $pq$
Lagrangian $D(p_T\|q)$ where $p_T(\mathbf{x}) =
\exp\bigl(-G(\mathbf{x})/T\bigr)/Z(T)$ with $Z(T)$ being the
normalization of the Boltzmann distribution. The $pq$ Lagrangian
is minimized by the the product of the marginals of the Boltzmann
distribution, i.e. $q_i(x_i) = \int d\mathbf{x}_{-i} \;
p_T(\mathbf{x})$. Another example of update rules without
Euclidean analogues are the iterative focusing update rules
described in \cite{wolp04g}. Iterative focusing updates are
intrinsically tied into the fact that we're minimizing (the
distribution setting) an expectation value.

A subset of update rules arising from $qp$ and $pq$ Lagrangians are
described in \cite{wolp04g}. In all cases, the updates may be written
as multiplicative updating of $q$. The following is a list of the
update ratios $r_{q,i}(x_i) \equiv q^{t+1}_i(x_i) / q^t_i(x_i)$ of
some of those rules.

In all of the rules in this list, $F_G$ is a probability distribution
over $x$ that never increases between two $x$'s if $G$ does (e.g., a
Boltzmann distribution in $G(x)$).  In addition, const is a scalar
that ensures the new distribution is properly normalized and $\alpha$
is a stepsize.\endnote{As a practical matter, both Nearest Newton and
gradient-based updating have to be modified in a particular step if
their step size is large enough so that they would otherwise take one
off the unit simplex. This changes the update ratio for that step.
See~\cite{wobi04a}.} Finally, ``iterative focusing'' is a technique
that repeats the following process: It takes a distribution
$\tilde{q}$ as input. It then produces as output a new distribution
that is $\tilde{q}$ ``focused'' more tightly about $x$'s with good
$G(x)$. That focused distribution becomes $\tilde{q}$ for the next
iteration. See~\cite{wost06}.

\vspace{1mm}

\noindent \textit{Gradient descent of $qp$ distance to ${F_G}$:}
\begin{equation}
1 - \alpha \Bigl\{ \frac{ {\mathbb{E}}_{q^t}(\ln F_G | x_i) + \ln
(q^t_i(x_i))}{q^t_i(x_i\ )} \Bigr\} -
\frac{\text{const}}{q^t_i(x_i)}
\end{equation}

\vspace{1mm}

\noindent \textit{Nearest Newton descent of $qp$ distance to
${F_G}$:}
\begin{equation}
1 - \alpha \bigl\{ {\mathbb{E}}_{q^t}(\ln F_G | x_i) +
{\mbox{ln}}(q^t_i(x_i)) \bigr\} - \text{const}
\end{equation}

\vspace{1mm}

\noindent \textit{Brouwer updating for $qp$ distance to ${F_G}$:}
\begin{equation}
\text{const} \times \frac{e^{{\mathbb{E}}_{q^t}(\ln F_G  |
x_i)}}{q^t_i(x_i)}  \label{eq:brouwer.update}
\end{equation}

\vspace{1mm}

\noindent \textit{Importance sampling minimization of $pq$
distance to $F_G$:}
\begin{equation}
\text{const} \times {\mathbb{E}}_{q^t} \biggl(\frac{F_G}{q^t} |
x_i\biggr) \label{eq:pq.min}
\end{equation}

\vspace{1mm}

\noindent \textit{Iterative focusing of ${\tilde{q}}$ with
focusing function ${F_G}$ using $qp$ distance and gradient
descent:}
\begin{eqnarray}
1 - \alpha \Bigl\{\frac{ {\mathbb{E}}_{q^t}(\ln F_G | x_i) + \ln
\frac{q^t_i(x_i)}{{\tilde{q}}_i(x_i)}}{q^t(x_i)} \Bigr\} - \frac{
\text{const}}{q^t(x_i)}
\end{eqnarray}

\vspace{1mm}

\noindent \textit{Iterative focusing of ${\tilde{q}}$ with
focusing function ${F_G}$ using $qp$ distance and Nearest Newton:}
\begin{equation}
1 - \alpha \Bigl\{ {\mathbb{E}}_{q^t}(\ln F_G | x_i) + \ln
\frac{q^t_i(x_i)}{{\tilde{q}}_i(x_i)} \Bigr\} - \text{const}
\end{equation}

\vspace{1mm}

\noindent \textit{Iterative focusing of ${\tilde{q}}$ with
focusing function ${F_G}$ using $qp$ distance and Brouwer
updating:}
\begin{equation}
\text{const} \times e^{{\mathbb{E}}_{q^t}(\ln F_G | x_i)} \times
\frac{{\tilde{q}}(x_i)}{q_i^t(x_i)} \label{eq:iter.eF.qp.brouwer}
\end{equation}

\vspace{1mm}

\noindent \textit{Iterative focusing of ${\tilde{q}}$ with
focusing function $F_G$ using $pq$ distance:}
\begin{equation}
\text{const} \times {\mathbb{E}}_{\tilde{q}}(F_G | x_i) \times
\frac{\tilde{q}(x_i)}{q_i^t(x_i)} \label{eq:iter.F.pq}
\end{equation}

\vspace{1mm}

\noindent Note that some of these update ratios are themselves proper
probability distributions, e.g., the Nearest Newton update ratio.

This list highlights the ability to go beyond conventional
Euclidean optimization update rules, and is an advantage of
embedding the original optimization problem in a problem over a
space of probability distributions. Another advantage is the fact
that the distribution itself provides much useful information
(e.g., sensitivity information). Yet another advantage is the
natural use of Monte Carlo techniques that arise with the
embedding, and allow the optimization to be used for adaptive
control.

\subsubsection{Relation of PC to other distribution-based optimization
algorithms}

There is some work on optimization that precedes PC and that has
directly considered the distribution $q$ as the object of
interest. Much of this work can be viewed as special cases of PC. In
particular deterministic annealing ~\cite{duha00} is ``bare-bones''
parallel Brouwer updating. This involves no data-aging (or any other
scheme to avoid thrashing of the agents), difference utilities,
etc..\endnote{Indeed, as conventionally cast, deterministic annealing
assumes the conditional $G$ can be evaluated in closed form, and
therefore has no concern for Monte Carlo sampling issues.}

More tantalizingly, the technique of probability
matching~\cite{sajo} uses Monte Carlo sampling to optimize a
functional of $q$. This work was in the context of a single agent,
and did not exploit techniques like data-ageing. Unfortunately,
this work was not pursued.

Other work has both viewed $q$ as the fundamental object of interest
and used techniques like data-aging and difference utilities. In
particular, this is the case with the COllective INtelligence (COIN)
work \cite{wotu99a,wotu03a,wolp02}. However this work was not based on
information-theoretic considerations and had no explicit objective
function for $q$. It was the introduction of such considerations
that resulted in PC.

Another interesting body of early work is the cross entryop (CE)
method~\cite{rukr04}. This work is the same as iterative focusing
using $pq$ distance. The CE method does not consider the formal
difficulty with iterative focusing identified in~\cite{wost06}, or
any of the potential solutions to that problem discussed there.
That difficulty means that even with no sampling error (i.e., if
all estimated quantities are in fact exactly correct), in general
there are no guarantees that the algorithm converges to an optimal
$x$.

Other early work grew out of the Genetic Algorithms community. This
work was initiated with MIMIC~\cite{deis97}, and has since developed
into the Estimation of Distribution Algorithms (EDA)
approach~\cite{lola05}. A number of important issues were raised in
this early work, e.g., the importance of information theoretic
concepts.

For the most part however, especially in its early stages (e.g., with
MIMIC), this work viewed the samples as the fundamental object of
interest, rather than view the distribution being sampled that
way. Little concern arises for what the objective function {\it{of the
distribution being sampled}} should be, and of how samples can be used
to achieve that optimization.

This means that there is little concern for issues like the (lack of)
convexity of that implicit objective function. This prevents EDA's
from fully exploiting the power of continuous space optimization,
e.g., the absence of local minima with convex problems~\cite{bova03}.
Similarly, it means that with EDA's there is no concern for cases
where the distribution objective function can be optimized in closed
form, without any need for sampling at all. Nor is there widespread
appreciation for how old sample $x$'s can be re-used to help guide the
optimization (as for example they are in PC's adaptive importance
sampling~\cite{wost06}).

This contrasts with PC, whose distinguishing feature is that it
does not treat the variable ${\mathbf{x}}$ as the fundamental
object to be optimized, but rather the distribution across it,
$q$.  So for example, in PC, samples of that distribution are only
used if necessary to estimate quantities that cannot be evaluated
other ways; the fundamental objective function is stated in terms
of $q$. Indeed, in the cases considered in this paper, as well as
earlier PC work like that reported in~\cite{mawo04a}, no such
samples of $q$ arise.

Shortly after the introduction of PC, a variant of its Monte
Carlo version of parallel Brouwer updating has been introduced,
called the MCE method \cite{rubi05}. In this variant the annealing
of the Lagrangian doesn't involve changing the temperature $\beta$,
but instead changing the value of a constraint specifying
$\mathbb{E}_q(G)$. Accordingly, rather than jump directly to the
($\beta$-specified) solution given above, one has to solve a set of
coupled nonlinear equations relating all the $q_i$. (Another
distinguishing feature is no data-ageing, difference utilities or
the like are used in the MCE method.) The MCE method has been
justified with the KL argument reviewed above rather than with
``ratchet''-based maximum entropy arguments. This has redrawn
attention to the role of the argument-ordering of the KL distance,
and how it relates Brouwer updating and the CE method.

Another body of work related to PC is (loopy) propagation algorithms,
Bethe approximations, and the like ~\cite{mack03}. These techniques
can be seen as an alternative to semicoordinate transformations for
how to go beyond product distributions. Unlike those approaches, we
are guaranteed to reach a local minimum of free energy. (If we were to
use $pq$ KL distance rather than $qp$ KL distance, we would get to a
$global$ minimum of free energy.) In addition, via utilities like AU
and WLU (see appendix), we can exploit variance reduction techniques
absent from those other techniques. Similarly those other techniques
do not make use of data-aging.

Finally, there is also work that has been done after the
introduction of both the CE method and PC that is closely related
to both. Such methods are typically sample based, and a good
summary of these approaches can be found in~\cite{muho05}. The use
of junction tree factorizations for $q$ utilized in the above
methods can be extended to PC theory, and results along these
lines will be presented elsewhere.

\section{Conclusion}
A distributed constrained optimization framework based on
probability collectives has been presented. Motivation for the
framework was drawn from an extension of full-rationality game
theory to bounded rational agents. An algorithm that is capable of
obtaining one or more solutions simultaneously was developed and
demonstrated on two problems. The results show a promising, highly
distributed, off-the-shelf approach to constrained optimization.

There are many avenues for future exploration. Alternatives to the
Lagrange multiplier method used here can be developed for constraint
satisfaction problems. By viewing the constraints as separate
objectives, a Pareto-like optimization procedure may be developed
whereby a gradient direction is chosen which is constrained so that
no constraints are worsened. This idea is motivated by the highly
successful WalkSAT \cite{skc93} algorithm for $k$-sat in which spins
are flipped only if no previously satisfied clause becomes
unsatisfied as a result of the change.

Probability collectives also offer promise in devising new methods
for escaping local minima. Unlike traditional optimization methods
where monotonic transformations of the objective leave local minima
unchanged, such transformations will alter the local minima
structure of the Lagrangian. This observation, and alternative
Lagrangians (see \cite{Rub01} for a related approach using a
different minimization criterion) offer new approaches for improved
optimization.

\vspace{2mm} \noindent {\bf{ACKNOWLEDGEMENTS}} \noindent We would
like to thank Charlie Strauss for illuminating conversation and Bill
Dunbar and George Judge for helpful comments on the manuscript.

\theendnotes

\appendix

\section{Statistical estimation to update $q$}
\label{sec:mc}

Using either of the update rules Eqs. \eqref{BrouwerEq} or
\eqref{nearNewtEq} requires knowing $\mathbb{E}_{q^t_{-
i}}\!(G|x_i)$, defined in Eq. \eqref{condExpEq}. However, often we
cannot efficiently calculate all the terms $\mathbb{E}_{q^t_{-
i}}\!(G|x_i)$. To use our update rules in such situations we can use
Monte Carlo sampling, as described in this section.

\subsection{Monte Carlo sampling}

In the Monte Carlo approach, at each timestep every agent $i$ samples
its distribution $q_i$ to get a point $x_i$. Since we have a product
distribution $q$, these samples provides us with a sample $\mathbf{x}$
of the full joint distribution $q$. By repeating this process $L$
times we get a ``block'' of such joint samples. The $G$ values in that
block can be used by each agent separately to estimate its updating
values $\mathbb{E}_{q^t_{- i}}\!(G|x_i)$, for example simply by
uniform averaging of the $G$ values in the samples associated with
each $x_i$. Note that the single set of samples can be used no matter
how many agents are in the system; we don't need a different Monte
Carlo process for each agent to estimate their separate
$\mathbb{E}_{q^t_{- i}}\!(G|x_i)$.

All agents (variables) sample moves (variable settings) independently,
and coupling occurs only in the updates of the $q_i$. As we have seen
this update (even to second order) for agent $i$ depends only on the
conditional expectations $\mathbb{E}_{q_{- i}}(G|x_i)$ where $q_{- i}$
describes the strategies used by the other agents. Thus, if we are
using Monte Carlo, then the only information which needs to be
communicated to each agent is the $G$ values upon which the estimate
will be based. Using these values each agent independently updates its
strategy (its $q_i$) in a way which collectively is guaranteed to
lower the Lagrangian.

If the expectation is evaluated analytically, the $i$th agent
needs the $q_j$ distributions for each of the $j$ agents involved
in factors in $G$ that also involve $i$. For objective functions
which consists of a sum of local interactions each of which
individually involves only a small subset of the variables (e.g.
the problems considered here), the number of agents that $i$ needs
to communicate with is typically much smaller than $n$.

\subsection{Difference utilities for faster Monte Carlo convergence}

The basic Monte Carlo approach outlined above can be slow to
converge in high-dimensional problems.  For the problems
considered in this paper this is irrelevant, since
$\mathbb{E}_{q^t_{-i}}\!(G|x_i)$ may be efficiently calculated in
closed form for all agents $i$ and their moves $x_i$, so we don't
need to use Monte Carlo sampling. For completeness though here we
present details of an improvement of the basic Monte Carlo
approach that converges far more quickly.

Scrutiny of the two update rules \eqref{BrouwerEq} and
\eqref{nearNewtEq} reveals that we don't actually require the
values $\mathbb{E}_{q^t_{- i}}\!(G|x_i)$ for all $x_i$. All that
is needed are the differences $\mathbb{E}_{q^t_{- i}}\!(G|x_i) -
\mathbb{E}_{q^t_{- i}}\!(G|x'_i)$ for pairs of distinct moves
$x_i$ and $x'_i$. (For both update rules the other degrees of
freedom in the values $\mathbb{E}_{q^t_{- i}}\!(G|x_i)$ are
absorbed into the factor in the update rule ensuring normalization
of $q_i$.) In other words, to update $q_i$ we can replace the
sample averages of $G(\mathbf{x})$ with the sample averages of the
associated {\it{difference utility}} $g^i(\mathbf{x}) =
G(\mathbf{x}) - D^i(\mathbf{x}_{- i})$ for any function
$D^i(\mathbf{x}_{- i})$. The differences in expectation values
$\mathbb{E}_{q^t_{- i}}\!(g^i|x_i) - \mathbb{E}_{q^t_{-
i}}\!(g^i|x'_i)$ are invariant with respect to choice of $D^i$.
However the choice of $D^i$ can have a major effect on the
statistical accuracy of our Monte Carlo estimation.

We exploit the freedom granted by the introduction of
$D^i(\mathbf{x}_{-i})$ to derive quantities which may be more
easily estimated by Monte Carlo methods.  In particular, we may
define $D^i$ by requiring that the maximum likelihood estimator of
the gradient has the lowest possible variance. In appendix
\ref{diffUtilApp} we show that the choice
\begin{equation}
D^i(\mathbf{x}_{-i}) = \biggl[ \sum_{x''_i}
\frac{1}{L_{x''_i}}\biggr]^{-1} \sum_{x'_i} \frac{G(x'_i,
\mathbf{x}_{-i})}{L_{x'_i}} .
\end{equation}
gives the lowest possible variance. In the above equation
$L_{x_i}$ is the numbers of times that the $i$th variable assumed
value $x_i$ in a block of $L$ Monte Carlo samples.

The associated difference utility, $g^i(\mathbf{x}) =
G({\mathbf{x}}) - D^i(\mathbf{x}_{-i})$, is called the
\textit{Aristocrat utility} (AU). An approximation to it was
investigated in \cite{wotu01a,wotu04,wotu02a} and references
therein. AU itself was derived in ~\cite{wolp03a}. AU minimizes
variance of gradient descent updating \textit{regardless of the
form of $q$}.

Sometimes not all the terms in the sum in AU can be stored,
because $|\mathcal{X}_i|$ and/or the block size is too large. That
sum can be approximated by replacing the $L_{x_i}$ values in the
definition of AU with $q(x_i)L$. This replacement also provides a
way to address cases where one or more $L_{x_i} =
0$.{\endnote{Another alternative is to use data-ageing so that
data from preceding Monte Carlo blocks can be re-used. Yet another
alternative is to force a sample for each $x_i$ value that has no
samples at the end of the block, $x'_i$. This is done by drawing a
random sample of the distribution $\delta(x_i -
x'_i)q_{-i}(\mathbf{x}_{-i})$.}}  Similarly, for computational
reasons it may be desirable to approximate the weighted average of
$G$ over all $x'_i$ which defines AU.

The sum over $x'_i$ occurring in AU should not be confused with
the sum over $x'_i$ in Eq. \eqref{etaEq} that pulls the gradient
estimate back into the unit simplex. The sum here is over values
of $G$ for counterfactual sample pairs $(x'_i, \mathbf{x}_{-i})$.
(The other sum is over values of our gradient estimate at all of
its arguments.) When the functional form of $G$ is known it is
often the case that there is cancellation which allows AU be
calculated directly, in one evaluation, an evaluation which can be
cheaper than that of $G(\mathbf{x})$.  When this is not the case
their evaluation incurs a computational cost in
general.{\endnote{To Monte Carlo estimate a conditional expected
utility for agent $i$ it is crucial that $\mathbf{x}_{-i}$ be
generated by sampling the associated distribution $q_{-i}$.
However the value $x_i$ can be set in any fashion whatsoever.}}
This cost is offset by the fact that those evaluations allow us to
determine the value of AU not just for the actual point $(x_i,
\mathbf{x}_{-i})$, but in fact for all points $\{(x_i',
\mathbf{x}_{-i}) | x'_i \in \mathcal{X}_i\}$.

Nonetheless, there will be cases where evaluating AU requires
evaluating all possible $G(x'_i, \mathbf{x}_{-i})$, and where the
cost of that is prohibitive, even if it allows us to update AU for
all $x_i$ at once.  Fortunately there are difference utilities
that are cheaper to evaluate than AU while still having less
variance than $G$. In particular, note that the weighting factor
$L_{x'_i}^{-1} / \sum_{x''_i}L_{x''_i}^{-1}$ in the formula for AU
is largest for those $x_i$ which occur infrequently, i.e. that
have low $q_i(x_i)$.  This observation leads to the
\emph{Wonderful Life Utility} (WLU), which is an approximation to
AU that (being a difference utility) also has zero bias:
\begin{equation}\label{eq:WLU}
\mbox {$ g^{WLU}_i(x_i, \mathbf{x}_{- i}) = G(x_i, \mathbf{x}_{-
i)}) - G(x^{\text{clamp}}_i, \mathbf{x}_{- i})$}.
\end{equation}
In this formula, $x^{\text{clamp}}_i = \argmin_{x_i} L_{x_i}$ or
if we wish to be more conservative, $\argmin_{x_i} q_i(x_i)$,
agent $i$'s lowest probability move
\cite{wolp03b,wolp04a}.\endnote{Just like AU, less sophisticated
versions of WLU were previously explored under the same name, with
examples again arising in ~\cite{wotu01a,wowh00} and references
therein.}

Given the derivation of AU, one would expect that in general WLU will
work best if $x^{\text{clamp}}_i$ is set to the $x$ that is least
likely. As the run unfolds, which $x$ has that property will
change. However changing $x^{\text{clamp}}_i$ accordingly will void
the assumptions underlying the derivation of AU. In practice then,
while one usually can update $x^{\text{clamp}}_i$ in such a dynamic
fashion occasionally, doing so too frequently can cause problems.

\subsection{Discussion of Monte Carlo sampling}

Note that the foregoing analysis breaks down if any of the
$L_{x_i} = 0$. More generally it may break down if just one or
more of the $q(x_i)$ are particularly small in comparison to the
others, even if no $L_{x_i}$ is exactly zero. The reason for this
is that our approximation of the average over $n_i^x$ (see
Appendix \ref{diffUtilApp}) with the average where no $L_{x_i} =
0$ breaks down. Doing the exact calculation with no such
approximation doesn't fix the situation
--- once we have to assign non-infinitesimal probability to
$L_{x_i} = 0$, we're allowing a situation in which the gradient
step would take us off the simplex of allowed $q \in \cal{Q}$. We
might try to compensate for this by reducing the stepsize, but in
general the foregoing analysis doesn't hold if stepsize is reduced
in some situations but not in any others. (Variable stepsize
constitutes a change to the update rule. Such a modification to
the update rule must be incorporated into the analysis --- which
obviates the derivation of AU.)

One way to address this scenario would be to simply zero out the
probability of agent $i$ making any move $x_i$ for which $q_i(x_i)$ is
particular small. In other words, we can redefine $i$'s move space to
exclude any moves if their probability ever gets sufficiently
small. This has the additional advantage of reducing the amount of
``noise'' that agents $j \ne i$ will see in the next Monte Carlo
block, since the effect of agent $i$ on the value of $G$ in that block
is more tightly constrained.

There several ways to extend the derivation of AU, which only
addresses estimation error for a single agent at a time, and for
just that agent's current update. One such extension is to have
agent $i$'s utility set to improve the accuracy of the update
estimation for agents $j \ne i$. For example, we could try to bias
$q_i$ to be peaked about only a few moves, thereby reducing the
amount of noise those other agents $j \ne i$ will see in the next
Monte Carlo block due to variability in $i$'s move choice. Another
extension is to have agent $i$'s utility set to improve the
accuracy of its estimate of its update for future Monte Carlo
blocks, even at the expense of accuracy for the current block.

Strictly speaking, the derivation of AU only applies to gradient
descent updating of $q$. Difference utilities are unbiased
estimators for the Nearest Newton update rule, so long as each $i$
estimates $\mathbb{E}_{q^t}(g^i)$ as $q^t_i(x_i)$ times the estimate
of $\mathbb{E}_{q^t}(g^i | x_i)$,
\begin{equation*}
\sum_{x_i} {\hat{g}}^i_{x_i}(n_i)  q^t_i(x_i)
\end{equation*}
rather than as the empirical average over all samples of
$g^i$,\endnote{To see this, say we add a function
$D^i(\mathbf{x}_{-i})$ to $G({\mathbf{x}})$ both places $G$ occurs
in the Nearest Newton update rule. Then since $q_i(x_i)$ is
evaluated exactly, the data average of the empirical estimates of
those two terms involving $D^i$ both give the expectation value
$\alpha^t q_i^t(x_i){\mathbb{E}}_{q^t_{-i}}(D^i)$ exactly.
Accordingly, those two terms cancel.}
\begin{equation*}
\sum_{x_i} {\hat{g}}^i_{x_i}(n_i)  \frac{L_{x_i}}{L} .
\end{equation*}
However the calculation for how to minimize the
variance must be modified. Redoing the algebra above, the analog of AU
for the Nearest Newton rule arises if we replace
\begin{equation}
\frac{1}{L_{x_i}} \rightarrow \frac{[q_i(x_i)]^2} {L_{x_i}}
\biggl\{[1 - q_i(x_i)]^2 + \sum_{x'_i \ne x_i}[q_i(x'_i)]^2
\biggr\}
\end{equation}
throughout the equation defining AU.  Similar considerations apply to
Brouwer updating as well.  Nonetheless, in practice AU and WLU as
defined above work well (and in particular far better than taking $g^i
= G$) for the other updating rules as well.

For gradient descent updating, minimizing expected quadratic error of
our estimator of $\mathbb{E}_{q^t_{- i}}\!(G|x_i)$ corresponds to
making a quadratic approximation to the Lagrangian surface, and then
minimizing the expected value of the Lagrangian after the gradient
step \cite{wolp03b}. More generally, and especially for other update rules,
some other kind of error measure might be preferable. Such measures
would differ from the bias-variance decomposition. We do not consider
such alternatives here.

Note that the agents are completely ``blind'' in the Monte Carlo
process outline above, getting no information from other agents
other than the values of $G({\mathbf{x}})$. When we allow some
information to be transmitted between the agents we can improve
the estimation of $\mathbb{E}_{q^t_{- i}}\!(G|x_i)$ beyond that of
the simple Monte Carlo process outlined above.  For example, say
that at every timestep the agent $i$ knows not just its own move
$x_i$, but in fact the joint move ${\mathbf{x}}$. Then as time
goes on it accumulates a training set of pairs \{$({\mathbf{x}},
G({\mathbf{x}}))$\}. These can be used with conventional
supervised learning algorithms \cite{duha00} to form a rough
estimate, ${\hat{G}}$, of the entire function $G$. Say that in
addition $i$ knows not its own distribution $q_i(x^t_i)$, but in
fact the entire joint distribution, $q({\mathbf{x}}^t)$. Then it
can use that joint distribution together with ${\hat{G}}$ to form
an estimate of $\mathbb{E}_{q^t_{- i}}\!(G|x_i)$. That estimate is
in addition to the one formed by the blind Monte Carlo process
outlined above. One can then combine these estimates to form one
superior to both. See \cite{lewo04b}.

Even when we are restricted to a blind Monte Carlo process, there
are many heuristics that when incorporated into the update rules
that can greatly improve their performance on real-world problems.
In this paper we examine problems for which joint distributions
$q$ are known to all agents as well as the function form of
$G({\mathbf{x}})$ and the required expectations
$\mathbb{E}_{q^t_{- i}}\!(G|x_i)$ may be obtained in closed form.
So there is no need for Monte Carlo approximations. Accordingly
there is no need for those heuristics, and there is not even any
need to using difference utilities. Empirical investigations of
the effects of using difference utility functions and the
heuristics may be found in ~\cite{biwo04a,biwo04b,biwo04c}.

\section{Utility Derivation} \label{diffUtilApp}

Say we are at a timestep $t$ at the end of a Monte Carlo block,
and consider the simplest updating rule. This is gradient descent
updating, in which we wish to update $q_i$ at a timestep $t$ by
having each agent $i$ take a small step in the direction (cf.
Eq.~\eqref{eq:nabla})
\begin{equation*}
\mathbf{f}^{i,G} \triangleq -\biggl[\frac{\partial
\mathcal{L}(q^t)}{\partial q_i(x_i^1)}, \cdots, \frac{\partial
\mathcal{L}(q^t)}{\partial q_i(x_i^{|\mathcal{X}_i|})} \biggr] -
\eta_i(q) \mathbf{1}
\end{equation*}
where $\eta_i(q)$ was defined in Eq.~\eqref{etaEq}, $\mathbf{1}$
is the vector of length $|\mathcal{X}_i|$ all of whose components
are 1, and $x_i^1, \cdots, x_i^{|\mathcal{X}_i|}$ are the
$|\mathcal{X}_i|$ moves available to agent $i$. In general, there
will be some error in $i$'s estimate of that step, since it has
limited information about $q_{-i}^t$. Presuming quadratic loss
reflects quality of the update, for agent $i$ the Bayes-optimal
estimate of its update is the posterior expectation
\begin{equation*}
\int dq^{t} \; P(q_{-i}^{t} \mid n_i) \, \mathbf{f}^{i,G}
\end{equation*}
where $n_i$ is all the prior knowledge and data that $i$ has, and
the dependence of $\mathbf{f}^{i,G}$ on $q_{-i}^{t}$ is
implicit.\endnote{Quadratic loss can be roughly related to the
assumption that the Lagrangian is locally well-approximated by a
paraboloid. We want to estimate $\mathbf{f}$ using an estimator
$\hat{\mathbf{f}}(n)$ which is a function of prior information
$n$. If we penalize errors quadratically (i.e. the error incurred
when the estimate is $\hat{\mathbf{f}}$ and the true value is
$\mathbf{f}$ is $\text{Err}(\mathbf{f}, \hat{\mathbf{f}}) =
\|\hat{\mathbf{f}}-\mathbf{f}\|^2$), then the expected error
conditioned on $n$ is $\mathbb{E}_{\mathbf{f}|n}(\text{Err}) =
\int d\mathbf{f} \, P(\mathbf{f}|n) \|\hat{\mathbf{f}} -
\mathbf{f}\|^2 = \|\hat{\mathbf{f}}\|^2 - 2 \hat{\mathbf{f}}^\top
\mathbb{E}_{\mathbf{f}|n}(\mathbf{f}) +
\mathbb{E}_{\mathbf{f}|n}(\|\mathbf{f}\|^2)$. Minimizing this with
respect to the estimator $\hat{\mathbf{f}}$ gives the optimal
quadratic loss estimator $\hat{\mathbf{f}}(n) =
\mathbb{E}_{\mathbf{f}|n}(\mathbf{f})$ Uncertainty in $\mathbf{f}$
arises due to uncertainty in $q_{-i}$, and hence we integrate over
the unknown $q$.} $P(q_{-i}^t|n_i)$ is a probability distribution
over likely values of $q_{-i}^t$ given the information $n_i$
available to agent $i$.

Now agent $i$ can evaluate $\ln q_i(x_i)$ for each of its moves
$x_i$ exactly. However to perform its update it still needs the
integrals
\begin{equation*}
\int dq^{t} \; P(q_{-i}^{t} | n_i) \, \mathbb{E}_{q^t_{- i}}\!(G |
x_i)
\end{equation*}
(recall Eq.~\eqref{gradLEq}). In general these integrals can be
very difficult to evaluate. As an alternative, we can replace
those integrals with simple maximum likelihood estimators of them,
i.e., we can use Monte Carlo sampling. In this case, the prior
information, $n_i$, available to the agent is a list,
$\mathfrak{L}$, of $L$ joint configurations $\mathbf{x}$ along
with their accompanying objective values $G(\mathbf{x})$.

To define this precisely, for any function $h(\mathbf{x})$, let
$\hat{\mathbf{h}}(n_i)$ be a vector of length $|\mathcal{X}_i|$
which is indexed by $x_i$. The $x_i$ component of
$\hat{\mathbf{h}}(n_i)$ is indicated as $\hat{h}_{x_i}(n_i)$. Each
of its components is given by the information in $n_i$. The
$x_i$'th such component is the empirical average of the values
that $h$ had in the $L_{x_i}$ samples from the just-completed
Monte Carlo block when agent $i$ made move $x_i$, i.e.
\begin{equation*}
\hat{h}_{x_i}(n_i) = \frac{1}{L_{x_i}} \sum_{\mathbf{x} \in
\mathfrak{L}_{x_i}} h(\mathbf{x})
\end{equation*}
where $\mathfrak{L}_{x_i}$ is the set of $\mathbf{x}$ in
$\mathfrak{L}$ whose $i$th component is equal to $x_i$, and where
$L_{x_i} = |\mathfrak{L}_{x_i}|$. Given this notation, we can
express the components of the gradient update step for agent $i$
under the simple maximum likelihood estimator as the values
\begin{equation}
\hat{f}^{i,G}_{x_i}(n_i) = -\{{\hat{G}}_{x_i}(n_i) + T \ln
q_i(x_i)\} \; -\; {\hat{\eta}}(n_i) \label{eq:emp_grad}
\end{equation}
where
\begin{equation}
\hat{\eta}_i(n_i) \triangleq \frac{-1}{|X_i|} \sum_{x'_i} \bigl \{
\hat{G}_{x'_i}(n_i) + T \ln q_i(x'_i) \bigr\}.
\end{equation}

Unfortunately, often in very large systems the convergence of
$\hat{G}(n_i)$ with growing $L$ is very slow, since the distribution
sampled by the Monte Carlo process to produce $n_i$ is very
broad. This suggests we use some alternative estimator.  Here we focus
on estimators that are still maximum likelihood, just with a different
choice of utility.  In particular, we will focus on estimators based
on difference utilities, briefly introduced in the preceding appendix.

To motivate such estimators more carefully, first posit that the
differences $\mathbb{E}_{q^t_{- i}}\!(G | x_i) -
\mathbb{E}_{q^t_{- i}}\! (G | x'_i)$, one for each $(x_i, x'_i)$
pair, are unchanged when one replaces $G$ with some other function
$g_i$. So the change is equivalent to adding a constant to $G$, as
far as those differences are concerned. This means that if agent
$i$ used $q_{-i}^{t}$ to evaluate its expectation values exactly,
then its associated update would be unchanged under this
replacement. (This is due to cancellation of the equivalent
additive constant with the change that arises in $\eta_i(n_i)$
under the replacement of $G(\mathbf{x})$ with $g^i(\mathbf{x})$).
It is straight-forward to verify that the set of all $g^i$
guaranteed to have this character, regardless of the form of $q$,
is the set of \textit{difference utilities}, $g^i({x}) = {G}({x})
- D^i(\mathbf{x}_{-i})$ for some function $D^i$. $G$ itself is the
trivial case $D^i(\mathbf{x}_{-i}) = 0 \; \forall x_i$.

On the other hand, if we use a difference utility rather than $G$
in our maximum likelihood estimator then it is the sample values
of $P(g^i)$ that generate $n_i$, and we use the associated
$x_i$-indexed vector $\hat{g}^i_{x_i}(n_i)$ rather than
$\hat{G}_{x_i}(n_i)$ to update each $q_i$.  For well-chosen $D^i$
it may typically be the case that $\hat{g}^i(n_i)$ has a far
smaller standard deviation than does $\hat{G}(n_i)$.  In
particular, if the number of coordinates coupled to $i$ through
${G}$ does not grow as the system does, often such difference
utility error bars will not grow much with system size, whereas
the error bars associated with $G$ will grow greatly. Another
advantage of difference utilities is that very often the Monte
Carlo values of a difference utility are far easier to evaluate
than are those of $G$, due to cancellation in subtracting $D^i$.

To make this more precise we can solve for the difference utility
with minimal error bar. First as a notational matter, extend the
definition of $\mathbf{f}^{i,G}$ by replacing $G$ with (arbitrary)
$h$ throughout, writing that extended version as
$\mathbf{f}^{i,h}$. Then assuming no $L_{x_i} = 0$, we are
interested in the $g^i$ minimizing the data-averaged quadratic
error,
\begin{equation}
\mathbb{E}_{q_{-i},n_i^{g^i}}(\text{Err}) = \int d{q_{-i}} \;
P(q_{-i}) \int dn_i^{g^i} \; P(n^{g^i}_i | n^{x}_i, q_{-i}, g^i)
\|\mathbf{f}^{i,g^i} - {\hat{\mathbf{f}}}^{i,g^i}(n_i)\|^2,
\label{eq:bpv}
\end{equation}
where $P(q_{-i})$ reflects any prior information we might have
concerning $q_{-i}$ (e.g., that it is likely that the current
$\mathbf{f}^{i,g^i}$ is close to that estimated for the previous
block of $L$ steps), and $n^{g^i}_i$ is the set of values of the
private utility contained in $n_i$. (The associated $x_i$ values,
$n_i^x$, are independent of $g^i$ and $q_{-i}$ and therefore for
our purposes can be treated as though they are fixed.)

Now the components of $\hat{\mathbf{f}}^{i,g^i}(n_i)$ (one for
each $x_i$) are not independent in general, being coupled via
${\hat{\eta}}_i(n_i)$. To get an integrand that involves only
independent variables, we must work with only one $x_i$ component
at a time. To that end, rewrite the data-averaged quadratic error
as
\begin{equation*}
\sum_{x_i} \int dq_{-i} \; P(q_{-i}) \int dn_i^{g^i} \;
P(n^{g^i}_i | n^{x}_i, q_{-i}, g^i) [f^{i,g^i}_{x_i} -
{\hat{f}}^{i,g^i}_{x_i}(n_i)]^2
\end{equation*}
where $f_{x_i}^{i,g^i}$ is the $q_i(x_i)$ component of
$\mathbf{f}^{i,g^i}$. Our results will hold for all $q_{-i}$, so
we ignore the outer integral and focus on
\begin{equation}
\sum_{x_i} \int dn_i^{g^i} \; P(n^{g^i}_i | n^{x}_i, q_{-i}, g^i)
[f^{i,g^i}_{x_i} - {\hat{f}}^{i,g^i}_{x_i}(n_i)]^2. \label{AUEq1}
\end{equation}
For any $x_i$ the inner integral can be decomposed with the famous
bias-variance decomposition into a sum of two terms
~\cite{duha00}.\endnote{The bias-variance decomposition decomposes
the error of an estimator into two contributions. Again let $f$ be
the true quantity and let $\hat{f}(n)$ be an estimator of $f$
formed from data $n$. We are interested in the error
$\text{Err}(\hat{f},f) = (\hat{f}-f)^2$ when averaged over
different data sets $n$ so we consider
$\mathbb{E}_{f,n}(\text{Err}) = \int dn df \, P(n,f) (\hat{f} -
f)^2$. By adding and subtracting $\mathbb{E}_{f}(f) -
\mathbb{E}_n(\hat{f})$ we have $\mathbb{E}_{f,n}(\text{Err}) =
\int dn df \, P(n,f)  \bigl( \{ \hat{f} - \mathbb{E}_n(\hat{f})\}
+ \{\mathbb{E}_{f}(f) - f\} + \{\mathbb{E}_n(\hat{f}) -
\mathbb{E}_{f}(f)\} \bigr)^2$. The cross term vanish when
integrating, and we are left with $\mathbb{E}_{f,n}(\text{Err}) =
\mathbb{E}_n\bigl( [\hat{f} - \mathbb{E}_n(\hat{f})]^2\bigr) +
\mathbb{E}_{f}\bigl( [f - \mathbb{E}_{f}(f)]^2\bigr) +
\bigl[\mathbb{E}_{f}(f) - \mathbb{E}_n(\hat{f})\bigr]^2$. The
first term is the variance of the estimator across different data
sets. The second term is the noise inherent in the quantity being
estimated which is independent of the estimator, and which we will
therefore ignore. The final term is the square of the bias of the
estimator.} The first of the two terms in our sum is the (square
of the) {\it{bias}}, $f^{i,g^i}_{x_i} -
\mathbb{E}_{n_i^{g^i}}(\hat{f}_{x_i}^{i,g^i})$, where
\begin{equation}
\mathbb{E}_{n_i^{g^i}} \bigl(\hat{f}^{i,g^i}_{x_i}(n_i^{g^i})
\bigr) \triangleq \int dn^{g^i}_i \; P(n^{g^i}_i | n^{x}_i,
q_{-i}, g^i) \hat{f}^{i,g^i}_{x_i}(n_i)
\end{equation}
is the expectation (over all possible sets of Monte Carlo sample
utility values $n^g_i$) of ${\hat{f}}^{i,g^i}_{x_i}(n_i)$. The
bias reflects the systematic trend of our sample-based estimate of
$f^{i,g^i}_{x_i}$ to differ from the actual $f^{i,g^i}_{x_i}$.
When bias is zero, we know that on average our estimator will
return the actual value it's estimating.

The second term in our sum is the \textit{variance},
\begin{equation}
\var\bigl(\hat{f}^{i,g^i}_{x_i}(n_i^{g^i})\bigr) \triangleq \int
dn^{g^i}_i \; P(n^{g^i}_i | n^{x}_i, q_{-i}, g^i) \; \bigl\{
\hat{f}^{i,g^i}_{x_i}(n_i) -
\mathbb{E}_{n_i^{g^i}}(\hat{f}^{i,g^i}_{x_i}) \bigr\}^2,
\end{equation}
In general variance reflects how much the value of our estimate
``bounces around'' if one were to resample our Monte Carlo block.
In our context, it reflects how much the private utility of agent
$i$ depends on its own move $x_i$ versus the moves of the other
agents. When $i$'s estimator is isolated from the moves of the
other agents ${\hat{f}}^{i,g^i}_{x_i}(n_i)$ is mostly independent
of the moves of the other agents, and therefore of $n_i$. This
means that variance is low, and there is a crisp
``signal-to-noise'' guiding $i$'s updates. In this situation the
agent can achieve a preset accuracy level in its updating with a
minimal total number of samples in the Monte Carlo block.

Plug the general form for a difference utility into the formula
for ${\hat{f}}^{i,g^i}_{x_i}(n_i)$ to see that (due to
cancellation with the ${\hat{\eta}}(n_i)$ term) its
$n^{g^i}_i$-averaged value is independent of $D^i$. Accordingly
bias must equal 0 for difference utilities. (In fact, difference
utilities are the only utility that is guaranteed to have zero
bias for all $q_{-i}$.) So our expected error reduces to the sum
over all $x_i$ of the variance for each $x_i$.

For each one of those variances again use Eq.~\ref{eq:emp_grad}
with $G$ replaced by $g^i$ throughout to expand
$\hat{f}^{i,g^i}_{x_i}(n_i)$. Since the $q_i(x_i)$ terms in that
expansion are all the same constant independent of $n_i$, they
don't contribute to the variance. Accordingly we have
\begin{eqnarray}
\var\bigl(\hat{f}^{i,g^i}_{x_i}(n_i^{g^i})\bigr) &=&
\var\biggl({\hat{g}}^i_{x_i}(n_i^g) - \frac{1}{|\mathcal{X}_i|}
\sum_{x'_i}{\hat{g}}^i_{x'_i}(n_i^g) \biggr)
\nonumber \\
&=& \var\biggl(\biggl[1 - \frac{1}{|\mathcal{X}_i|}\biggr]
{\hat{g}}^i_{x_i}(n_i^g) - \frac{1}{|\mathcal{X}_i|} \sum_{x'_i
\ne x_i} {\hat{g}}^i_{x'_i}(n_i^g)  \biggr).
\end{eqnarray}

Since $n^x_i$ is fixed and we are doing IID sampling, the two
expressions inside the variance function are statistically
independent.  In addition, the variance of a difference of
independent variables is the sum of the variances. Accordingly,
the sum over all $x_i$ of our variances is (cf. Eq.~\eqref{AUEq1})
\begin{eqnarray}
\sum_{x_i} \var\bigl(\hat{f}^{i,g^i}_{x_i}(n_i^g)\bigr) &=&
\sum_{x_i} \var\biggl(\biggl[1 - \frac{1}{|\mathcal{X}_i|}\biggr]
{\hat{g}}^i_{x_i}(n_i^g) - \frac{1}{|\mathcal{X}_i|} \sum_{x'_i
\ne x_i} {\hat{g}}^i_{x'_i}(n_i^g)  \biggr)
\nonumber \\
&=& \sum_{x_i} \biggl\{ \biggl[1 -
\frac{1}{|\mathcal{X}_i|}\biggr]^2
\var\bigl({\hat{g}}^i_{x_i}(n_i^g)\bigr) +
\frac{1}{|\mathcal{X}_i|^2} \sum_{x'_i \ne x_i}
\var\bigl({\hat{g}}^i_{x'_i}(n_i^g)\bigr) \biggr\} \nonumber \\
&=& \sum_{x_i} \biggl\{\biggl[1 -
\frac{1}{|\mathcal{X}_i|}\biggr]^2 + \frac{|\mathcal{X}_i| -
1}{|\mathcal{X}_i|^2} \biggr\} \var({\hat{g}}^i_{x_i}(n_i^g))
\nonumber \\
&=&  \frac{|\mathcal{X}_i| - 1}{|\mathcal{X}_i|} \sum_{x_i}
\var({\hat{g}}^i_{x_i}(n_i^g)) ,
\end{eqnarray}
where the third equation follows from the second by using the
trivial identity $\sum_a \sum_{b \ne a} F(b) = \sum_a F(a) \sum_{b
\ne a} 1$ for any function $F$.

Since for each such $x_i$ we are doing $L_{x_i}$-fold IID sampling
of an associated fixed distribution, the variance for each
separate $x_i$ is of the form $$ \int d{\mathbf{y}} \;
P({\mathbf{y}})  \biggl[ \frac{1}{L_{x_i}} \sum_{j=1}^{L_{x_i}}
y_j - \mathbb{E}_P\biggl(\frac{1}{L_{x_i}} \sum_{j=1}^{L_{x_i}}
y_j\biggr) \biggr ]^2
$$
for a fixed distribution $P(y_1, y_2, \ldots, y_{_{L_{x_i}}}) =
\prod_{j=1}^{L_{x_i}} P(y_j)$. We can again use the decomposition
of a variance of a sum into a sum of variances to evaluate this.
With the distribution $q^t$ implicit, define the single sample
variance for the value of any function $H(x)$, for move $x_i$, as
\begin{equation}
\var(H(x_i)) \triangleq \mathbb{E}([H]^2 | x_i) - [\mathbb{E}(H |
x_i)]^2.
\end{equation}
This gives
\begin{eqnarray}
\var(\hat{g}^i_{x_i}(n_i^g)) &=& \var(g^i(x_i)) \;/\; L_{x_i}
\end{eqnarray}
Collecting terms, we get
\begin{eqnarray}
\sum_{x_i} \var(\hat{f}^{i,g^i}_{x_i}(n_i^g)) &=&
\frac{|{\mathcal{X}}_i| - 1}{|{\mathcal{X}}_i|} \sum_{x_i}
\frac{\var(g^i(x_i))}{L_{x_i}}. \label{eq:diff_var}
\end{eqnarray}

Now $\var(A(\tau)) = (1/2) \sum_{t_{1},t_{2}}
P(t_{1})P(t_{2})[A(t_{1}) - A(t_{2})]^{2}$ for any random variable
$\tau$ with distribution $P(t)$. Use this to rewrite the sum in
Eq.~\eqref{eq:diff_var} as
\begin{equation*}
\frac{|{\mathcal{X}}_i| - 1}{2|{\mathcal{X}}_i|} \sum_{x_i}
L_{x_i}^{-1} \sum_{\mathbf{x}'_{_{-i}}, \mathbf{x}''_{_{-i}}}
q_{-i}(\mathbf{x}'_{-i}) q_{-i}(\mathbf{x}''_{-i}) [g^i(x_i,
\mathbf{x}'_{-i}) - g^i(x_i, \mathbf{x}''_{-i})]^{2}
\end{equation*}
Bring the sum over $x_i$ inside the other integrals, expand $g^i$,
and drop the overall multiplicative constant to get
\begin{eqnarray*}
\var(\hat{F}^i_{g^i}(n_i^g)) &\propto& \sum_{\mathbf{x}'_{_{-i}},
\mathbf{x}''_{_{-i}}}
q_{-i}(\mathbf{x}'_{-i}) q_{-i}(\mathbf{x}''_{-i})  \\
&& \;\;\;\;\;\;\; \sum_{x_i}  \frac{\bigl[G(x_i, \mathbf{x}'_{-i})
- G(x_i, \mathbf{x}''_{-i}) - \{D^i(\mathbf{x}'_{-i}) -
D^i(\mathbf{x}''_{-i})\}\bigr]^{2}}{L_{x_i}}.
\end{eqnarray*}
For each $\mathbf{x}'_{-i}$ and $\mathbf{x}''_{-i}$, our choice of
$D^i$ minimizes the sum so long as the difference in the curly
brackets obeys
\begin{equation*}
D^i(\mathbf{x}'_{-i}) - D^i(\mathbf{x}''_{-i}) = \sum_{x_i}
\frac{[G(x_i, \mathbf{x}'_{-i}) - G(x_i,
\mathbf{x}''_{-i})]}{L_{x_i}} \; / \; \sum_{x'_i}
\frac{1}{L_{x'_i}}.
\end{equation*}
This can be assured by picking
\begin{equation}
D^i(\mathbf{x}_{-i}) = \biggl[ \sum_{x'_i}
\frac{1}{L_{x'_i}}\biggr]^{-1} \sum_{x'_i} \frac{G(x'_i,
\mathbf{x}_{-i})}{L_{x'_i}} .
\end{equation}
for all $\mathbf{x}_{-i}$.

Note that AU minimizes variance of gradient descent updating
\textit{regardless of the form of $q$}. Indeed, being independent
of $q_{_{-i}}$, it minimizes our original $q_{_{-i}}$ integral in
Eq. ~\eqref{eq:bpv}, regardless of the prior $P(q_{_{-i}})$. For
the same reason it is optimal if the integral is replaced by a
worst-case bound over $q_{_{-i}}$.

\end{document}